%% file: main.tex
\documentclass[sigconf]{acmart}

\input{preamble/packages}

\input{preamble/math}

\copyrightyear{2023}
\acmYear{2023}
\setcopyright{acmlicensed}
\acmConference[ICAIF '23]{4th ACM International
Conference on AI in Finance}{November 27--29, 2023}{Brooklyn, NY, USA}
\acmBooktitle{4th ACM International Conference on AI in Finance (ICAIF '23),
November 27--29, 2023, Brooklyn, NY, USA}
\acmPrice{15.00}
\acmDOI{10.1145/3604237.3626841}
\acmISBN{979-8-4007-0240-2/23/11}

\AtBeginDocument{%
  \providecommand\BibTeX{{%
    \normalfont B\kern-0.5em{\scshape i\kern-0.25em b}\kern-0.8em\TeX}}}

\begin{document}

\title{SigFormer: Signature Transformers for Deep Hedging}


\author{Anh Tong}
\email{anhtong@kaist.ac.kr}
\affiliation{%
  \institution{KAIST}
  \country{South Korea}
}

\author{Thanh Nguyen-Tang}
\email{nguyent@cs.jhu.edu}
\affiliation{%
  \institution{Johns Hopkins University}
  \country{USA}
}

\author{Dongeun Lee}
\email{dongeun.lee@tamuc.edu}
\affiliation{%
  \institution{Texas A\&M University-Commerce}
  \country{USA}
}

\author{Toan Tran}
\email{v.toantm3@vinai.io}
\affiliation{%
  \institution{VinAI Research}
  \country{Vietnam}
}

\author{Jaesik Choi}
\email{jaesik.choi@kaist.ac.kr}
\affiliation{%
  \institution{KAIST/INEEJI}
  \country{South Korea}
}

\begin{abstract}  
  Deep hedging is a promising direction in quantitative finance, incorporating models and techniques from deep learning research. While giving excellent hedging strategies, models inherently requires careful treatment in designing architectures for neural networks. To mitigate such difficulties, we introduce \sigformer, a novel deep learning model that combines the power of path signatures and transformers to handle sequential data, particularly in cases with irregularities. Path signatures effectively capture complex data patterns, while transformers provide superior sequential attention.
  Our proposed model is empirically compared to existing methods on synthetic data, showcasing faster learning and enhanced robustness, especially in the presence of irregular underlying price data. Additionally, we validate our model performance through a real-world backtest on hedging the S\&P 500 index, demonstrating positive outcomes.

\end{abstract}

\maketitle

\input{1-intro}

\input{2-background}

\input{3-model}

\input{4-experiment}

\input{5-conclusion}

\begin{acks}
    We gratefully acknowledge the support received for this work, including the Institute of Information \& communications Technology Planning \& Evaluation (IITP) grant funded by Korean goverment (MSIT) under No. 2022-0-00984, which pertains to research in Artificial Intelligence, Explainability, Personalization, Plug and Play, Universal Explanation Platform. Additionally, we acknowledge support from the Artificial Intelligence Graduate School Program (KAIST) under No. 2019-0-00075, and from the Development and Study of AI Technologies to Inexpensively Conform to Evolving Policy on Ethics under No. 2022-0-00184.

We thank anonymous reviewers for their insightful feedback. We would like to extend our appreciation to Enver Menadjiev, Artyom Stitsyuk, and Hyukdong Kim for involvement in the early stage of this project. 
\end{acks}

\bibliographystyle{ACM-Reference-Format}
\bibliography{ref}

\appendix
\input{6-appendix}

\end{document}

%% file: preamble/packages.tex
\usepackage{url}
\usepackage{array}
\usepackage{amsmath}
\usepackage{amsfonts}
\usepackage{amsthm}
\usepackage{booktabs}
\usepackage{relsize}
\usepackage{nicefrac}
\usepackage{graphicx}
\usepackage{rotating}
\usepackage{nth}
\usepackage{soul}
\usepackage{bm}
\usepackage{mathtools}
\usepackage{wrapfig}

\usepackage{footnote}
\usepackage{multirow}
\usepackage{listings}

\usepackage{tikz}
\usepackage{namedtensor}
\usepackage{xspace}

\usepackage{xcolor}

\definecolor{codegreen}{rgb}{0,0.6,0}
\definecolor{codegray}{rgb}{0.5,0.5,0.5}
\definecolor{codepurple}{rgb}{0.58,0,0.82}
\definecolor{backcolour}{rgb}{0.95,0.95,0.92}

\lstdefinestyle{mystyle}{
    backgroundcolor=\color{backcolour},   
    commentstyle=\color{codegreen},
    keywordstyle=\color{magenta},
    numberstyle=\tiny\color{codegray},
    stringstyle=\color{codepurple},
    basicstyle=\ttfamily\footnotesize,
    breakatwhitespace=false,         
    breaklines=true,                 
    captionpos=b,                    
    keepspaces=true,                 
    numbers=left,                    
    numbersep=5pt,                  
    showspaces=false,                
    showstringspaces=false,
    showtabs=false,                  
    tabsize=2
}

%% file: preamble/math.tex
\newcommand{\expect}{\mathbb{E}}

\newcommand{\TR}{T((\R^d))}

\def\R{\mathbb{R}}

\newcommand{\iprod}[2]{\langle #1, #2\rangle} 

\DeclareMathOperator{\Sig}{Sig}

\newtheorem{prop}{Proposition}

\newtheorem{thm}{Theorem}

\theoremstyle{definition}
\newtheorem{definition}[thm]{Definition}

\newtheorem{rem}{Remark}

\DeclareMathOperator{\attn}{attn}
\DeclareMathOperator{\attention}{Attention}
\DeclareMathOperator{\attentionlayer}{AttentionLayer}
\DeclareMathOperator{\multihead}{MultiHead}
\DeclareMathOperator{\concat}{Concat}
\DeclareMathOperator{\head}{head}
\DeclareMathOperator{\softmax}{softmax}

\newcommand{\NN}{\textsc{nn}}
\newcommand{\sigformer}{\textsc{SigFormer}\xspace}
\newcommand{\rnn}{\textsc{rnn}\xspace}

%% file: 1-intro.tex
\section{Introduction}

The effective hedging of derivatives represents a crucial challenge in the field of mathematical finance. Over the years, various well-established approaches have been developed to derive tractable solutions for these problems. Among the conventional methods, classical risk management involves computing quantities known as ``greeks.'' Nevertheless, this approach heavily depends on somewhat unrealistic assumptions and settings, which may limit its applicability in certain scenarios. 

In recent times, there has been a paradigm shift in addressing this problem. Departing from the traditional techniques, \citet{deep_hedging} introduced a novel direction by harnessing the power of deep learning models. This innovative approach has the remarkable ability to handle a broader range of settings, even extending to high-dimensional cases, where conventional methods may fall short. By leveraging the capabilities of deep learning, the proposed framework opens up new possibilities for enhancing hedging strategies and exploring more realistic and adaptable solutions. 

Hedging a derivative involves making decisions to buy or sell the underlying instrument at specific time points to achieve risk-neutrality. Typically, it requires making an assumption regarding underlying price processes like the Hull-White model model~\cite{hull_white} and the Heston models~\cite{heston1993closed}. However, recent work~\cite{gatheral2014volatility} suggests that fractional stochastic volatility (FSV) models offer a more suitable choice, exhibiting properties that closely resemble those observed in financial markets. Modeling such a process involves the use of fractional Brownian models introduced in~\cite{mandelbrot1968fractional}.

Deep hedging~\cite{deep_hedging, deep_hedging_rough_vol,zhu2023stochastic} using recurrent neural networks (RNNs) might encounter challenges when dealing with irregular sequences from FSV modes. Some evidence from~\cite{deep_signature_transform} indicates that the RNN approach is less effective in this context. To handle the irregularity present in data, we propose 
incorporating path signatures. Path signatures~\cite{Lyons1998, Lyons2014RoughPS, lyons2023signature} are mathematical transformations capable of extracting features that describe the curves of paths as a collection of terms, potentially infinite in number. By employing these path signatures, we can effectively represent and capture the complexities of irregular data patterns~\cite{friz2020course}.

Additionally, we further leverage the ability to extract important features from signatures by incorporating transformers~\cite{attention_is_all_you_need}. Unlike the previous approaches like~\cite{deep_hedging, deep_hedging_rough_vol} that directly use market information as input into models, our approach takes signature outputs as input for transformers. While signatures handle roughness in data, transformers provide better sequential attention. Transformers are preferable 
 over RNNs in a vast number of machine learning applications due to the scalability in training and the capability to model long sequences. 
To the best of our knowledge, our work is the first to explore the ability of transformers with a deep hedging framework.



Our proposed model offers a distinct combination of signature computations and transformers. In contrast to a naive approach of directly applying transformers to signatures, we propose to incorporate multiple attention blocks designed to target selective and specialized terms of signature. Our novel design, driven by the fact that individual term in signatures possesses unique geometric properties, leads to a strong ability to handle the irregularity in the data, as we will show in our experiments. 


The paper makes the following key contributions: (1) introducing a novel deep learning model, named \sigformer, which carefully leverages the power of signatures and transformers in a principled manner, offering a novel and effective approach for sequential data analysis; (2) conducting an extensive empirical comparison of our proposed model with existing methods on various synthetic data settings. The results demonstrate that \sigformer exhibits faster learning curves and enhanced robustness, particularly in cases where underlying price data exhibit greater irregularity; (3) providing a backtest on real-world data for hedging the S\&P 500 (Standard \& Poor 500) index, validating the performance of our model and showcasing positive outcomes.


%% file: 2-background.tex
\section{Background}

This section offers a brief review of signatures in rough path theory and provides the background on deep hedging models and transformers.

\subsection{Signatures}
Here, we follow the standard notion in~\cite{Lyons2014RoughPS}.  Consider an extended tensor algebra
$$\TR := \{(a_0, a_1, \dots, a_i, \dots)| a_i \in (\R^d)^{\otimes n}\}.$$
Here, $(\R^d)^{\otimes n} = \underbrace{\R^d \otimes \cdots \otimes \R^d}_{n \text{ times}}$, and $(\R^d)^{0} \coloneqq \R$.
\begin{definition}[Signature]
Let $X: [0, T] \to \R^d$ be a continuous path.
The signature of $X$ over an interval $[s,t] \subset [0,T]$ is defined as an infinite series of tensors indexed by the signature order $n \in \mathbb{N}$,
\begin{equation}
    \Sig_{s,t}(X) \coloneqq \left(1, \Sig_{s,t}^1(X), \Sig_{s,t}^2(X),\dots,  \Sig^n_{s, t}(X), \dots \right) \in \TR,
\end{equation}
where
\begin{equation}
    \Sig^n_{s,t}(X) := \int_{s < u_1 < \dots < u_n < t} dX_{u_1} \otimes \dots \otimes dX_{u_n} \in (\R^d)^{\otimes n}.
\end{equation}
\end{definition}

In the lens of studying controlled differential equations,~\cite{Lyons2014RoughPS} emphasizes that signatures are the key tools in modeling non-linear systems with highly oscillatory signals. Many financial instruments are known to be one of such highly irregular time series. Intuitively, we can understand $\Sig^n(X)$ as the $n$-th moment tensor of the infinitesimal change given $n$ time points sampled uniformly on the path. Some references including~\cite{primer_sig_ml} give a comprehensive introduction of signatures for machine learning. 

\begin{prop}[Invariance to time parameterization]
Let $X: [0, T] \to \R^d$ be a continuous path. For any function $\varphi: [0,T] \to [0,T]$ that is continuously differentiable, increasing, and surjective, we have
$$\Sig(X) = \Sig(X \odot \varphi),$$
where $[X \odot \varphi](\cdot) = X(\varphi(\cdot))$.
\end{prop}

Proposition 1 indicates that the signature is invariant under reparameterization (see~\cite[Proposition 7.10]{friz2010multidimensional} for the proof). One may refer to~\cite[Figure 1]{signature_kernel_2} for an intuitive example of this property. Such an invariance is important in machine learning models to tackle symmetries in data, e.g., $SO(3)$ invariance in computer vision.

As noted in~\cite{deep_signature_transform}, signature transforms share a resemblance with the Fourier transform in the sense that a signal can be approximated well given a finite basis. Consider a non-linear function of path $X \to f(X)$, we can have a universal approximation of $f$ via a linear function of the signatures which is $f(X) \approx \iprod{W}{\Sig(X)}$, where $W$ is a linear weight. Such a property allows~\cite{sig_sde} calibrating financial models efficiently and provides the theory for optimal hedging~\cite{nonparam_pricing}.

\begin{prop}[Universal Nonlinearity]
\label{prop:universal_nonlinearity}
    Let $\mathcal{V}_{1}([0, T]; \R^d)$ be the space of continuous paths from some interval $[0, T]$ to $\R^d$. Suppose $\mathcal{K} \in \mathcal{V}_{1}([0, T]; \R^d)$ is compact and $f: \mathcal{K} \mapsto \R$ is continuous. 
    For any $\varepsilon > 0$ there exists a truncation level $n \in \mathbb{N}$ and coefficients $\alpha_i (\mathbf{J}) \in \R$ such that for every $X \in \mathcal{K}$, we have
    \begin{equation}
        \left\lvert f(X) - \sum_{i=0}^n \sum_{\mathbf{J} \in \{1, \dots, d\}^i} \alpha_i(\mathbf{J}) \Sig_{a, b}(X) \right\rvert \leq \varepsilon. 
    \end{equation}
\end{prop}
Proving this proposition involves showing signatures span an algebra and subsequently applying the Stone-Weierstrass theorem (refer to \cite[Theorem 1]{signature_kernel_1} for a comprehensive proof).

The following proposition is helpful in handling stream data.
\begin{prop}[Chen's identity~\cite{Lyons1998}]
\label{prop:chen}
Let $x, y: [a, b]\to \R^d$ be two continuous paths such that $x(a) = y(b)$. The concatenation
$x \star y$ yields the signature
\begin{equation}
    \Sig(x \star y) = \Sig(x) \otimes \Sig(y).
\end{equation}
\end{prop}

With the necessary background on signatures above, we will now delve into their practical implementations. In practice, the truncated version of the signature, denoted as $\Sig_{s,t}(X) = (\Sig_{s,t}^n(X))_{n=0,\dots, N}$, proves to be sufficiently expressive.

Several libraries offer implementations for signature computations, such as \texttt{esig}\footnote{\url{https://pypi.org/project/esig/}}, \texttt{iisignature}\cite{iisignature}, \texttt{signatory}\cite{kidger2021signatory}, and \texttt{signax}\footnote{https://pypi.org/project/signax/}.

\subsection{Deep Hedging}

Recently, deep hedging models~\cite{deep_hedging} emerged as a new paradigm for pricing and hedging models. Several works, such as~\cite{buehler2022lecture, buehler2022lecture_2, buehler2022deep, buehler2023deep} have further extended or integrated deep hedging in their research. 

\paragraph{Settings} Let us consider a market with a time horizon denoted by $T$. Trading is exercised at dates $0 = t_0 < t_1 <\cdots<t_n=T$, and $I_k$ represents the market information at each date $t_k$. In this market, there are $d$ hedging instruments, given by $S \coloneqq (S_k)_{k=0,\dots, n}$ where $S_k \in \R^d$. The liability of our derivative at $T$ is defined as $Z$. A hedge strategy is denoted as $\delta \coloneqq (\delta_k)_{k=0, \dots, n-1}$, with $\delta_k \in \R^d$.

Deep hedging, introduced by~\citet{buehler2022deep}, aims to find the optimal $\delta$ that minimizes the following objective under a pricing measure $\mathbb{Q}$ of financial market:
\begin{equation}
    \inf_\delta \expect_{\mathbb{Q}}[\rho(-Z + (\delta \cdot S)_T - C_T(\delta))],
    \label{eq:objective}
\end{equation}
where $\rho: \R \mapsto \R$ represents a convex risk measure~\cite{convex_risk_measure, ilhan2009optimal}, and $(\delta \cdot S)_T \coloneqq \sum_{k=0}^{n-1}\delta_k (S_{k+1} - S_{k})$ indicates the wealth at time $T$ resulting from the chosen hedge strategy. The term $C_T(\delta)$ denotes the trading cost incurred by $\delta$. For simplicity, we do not include the cost in this work and set $C_T(\delta)=0$.



\paragraph{Neural network architecture} The neural network architecture, proposed by~\cite{deep_hedging} to approximate $\delta$ is denoted as $\delta_{k}^{\theta}\coloneqq F^{\theta}\left(I_{k}, \delta_{k-1}^{\theta}\right)$. Here $F^{\theta}$ is a feed-forward neural network known for its universal approximation properties~\cite{hornik1989multilayer}. Notably, in addition to incorporating market information $I_k$, the model exhibits recurrent behaviors since it takes the previous hedge strategy $\delta^{\theta}_{k-1}$ as an input.

In detail, the architecture of the model consists of an input layer with $2d$ nodes, two hidden layers, each comprising $d+15$ nodes, and an output layer with $d$ nodes. The activation function used is $\sigma(x)=\max(x, 0)$, commonly known as the ReLU (Rectified Linear Unit) activation function.

The use of this architecture allows the neural network to effectively approximate the optimal hedging strategy $\delta$, considering both the historical hedging decisions and the evolving market information. This makes the model capable of capturing complex dependencies and patterns.

\subsection{Attention and Transformer}
\label{sec:attention}
The architecture of transformers~\cite{attention_is_all_you_need} is grounded in three key components: self-attention, multi-head attention, and feed-forward neural network. These essential elements are briefly presented here.

\paragraph{Self-attention} Self-attention is specified by three components: a query, a key, and a value. Let $X \in \R^{n \times d_x}$ represent the input data, and let $W_q \in \R^{d_{\attn} \times d_x}$, $W_k \in \R^{d_{\attn} \times d_x}$, and $W_v \in \R^{d_{\attn} \times d_x}$ denote the linear projections for query, key, and value, respectively. We define $Q = X W_q^\top$, $K = X W_k^\top$, and $V = X W_v^\top$. Then
\begin{equation}
    \attention(Q, K, V) \coloneqq \softmax \left(\frac{QK^\top}{\sqrt{d_x}}\right) V.
    \label{eq:self_attention}
\end{equation}
Intuitively, self-attention can be interpreted as an operation that encodes the process of determining the locations in a sequence that necessitate attention. Let us consider a sequence $X$ represented by elements $x_1, x_2, \dots, x_n$. The attention mechanism is established through the creation of a query ($q_t = x_t W_q^\top$), which is subsequently compared to other keys ($k_{\tau} = x_\tau W_k^\top$). The comparison is facilitated by a kernel function denoted as $\kappa(q_t, k_\tau)=\frac{\exp(q_tk_\tau^\top)}{\sum_s \exp(q_tk_s^\top)}$, serving to quantify the similarity between queries and keys. Finally, the element at position $t$ within the sequence is updated according to the aggregation $\sum_{\tau=1}^n \kappa(q_t, k_\tau) v_\tau$ with $v_\tau = x_\tau W_v^\top$. The value of $\kappa(q_t, k_\tau)$ determines the degree of attention the model directs towards $v_\tau$: a higher value of $\kappa(q_t, k_\tau)$ implies a stronger focus by the model on $v_\tau$.

\paragraph{Multi-head Attention}
Multi-head attention is an operation that concatenates various versions of attention and projects them into an appropriate space. It can be represented as
\begin{align*}
    \multihead \coloneqq & \concat(\head_1, \dots, \head_h) W_o, \\
    \text{where } \head_i \coloneqq & \attention(Q_i, K_i, W_i).
\end{align*}
Here, $W_o$ denotes the weight of the output projection, and $Q_i, K_i, V_i$ represents the query, key, value associated with weights $W_q^i, W_k^i, W_v^i$, respectively.

The multi-head attention mechanism facilitates the synthesis of joined representations and encodes richer information. It proves beneficial in 
 alleviating the sparsity inherent in the attention operation described in Eq.~\eqref{eq:self_attention} which arises from the softmax function.

\paragraph{Feed-forward network}
The feed-forward network (FFN), constituting the final component within transformers, consists of two linear layers and a ReLU activation. Notably, practical implementations often adopt the Gaussian Error Linear Unit (GELU) as the default choice. The FFN component can be expressed as follows:
\begin{equation}
    \textrm{FFN}(x) = \textrm{GELU}(xW_1^\top + b_1) W_2^\top + b_2.
\end{equation}
Here, $W_1$ and $W_2$ represent weight matrices; $b_1$ and $b_2$ denote corresponding bias vectors.

The signification of FFN has been extensively examined in the literature, wherein it is elucidated as a reservoir for information memory that facilitates the emergence capacities in large transformers~\cite{geva2021transformer}.

In addition to the above description of the essential components within the transformer, it is worth highlighting that other details, such as layer normalization and positional encoding, are elaborated in the original paper~\cite{attention_is_all_you_need}.

\section{Related Work}

\paragraph{Deep hedging} The concept of deep hedging was first introduced in the literature by~\citet{deep_hedging}. Since its inception, numerous research efforts have been devoted to extending and enhancing the deep hedging framework in various dimensions. For instance,~\citet{buehler2022deep} addressed the issue of drift removal, and recently explored incorporating reinforcement learning techniques~\cite{buehler2023deep}. Other relevant work, presented in~\citet{deep_hedging_rough_vol, zhu2023stochastic}, shares a similar goal aiming to enhance the deep hedging technique by using RNN architectures.

In our study, we propose a novel neural network architecture that combines two essential methodologies, namely signatures and transformers, to improve the deep hedging framework. In related work,~\citet{robust_hedging_GANs} extended the optimization objective by considering data uncertainty through an adversarial approach. It is worth noting that our proposed model directly employs signatures for learning data representations, whereas~\citet{robust_hedging_GANs} incorporates signatures as a regularization component.

With these advancements, our paper contributes to the refinement and enrichment of the deep hedging methodology by introducing a novel neural network architecture that integrates signatures and transformers. Moreover, we demonstrate the effectiveness of our proposed model in handling data uncertainty, which is also a crucial aspect in hedging strategies.


\paragraph{Applications of signatures} 
Signatures are a powerful mathematical tool utilized for modeling sequential data, as evidenced by several notable works~\cite{Lyons2014RoughPS, Lyons1998, lyons2023signature}. From theoretical standpoint, signatures hold a crucial role in the realm of rough path theory, establishing the fundamental basis for stochastic partial differential equations~\cite{friz2020course}.

The financial domain has witnessed an extensive array of applications of signatures, exemplified by the work of~\cite{sig_sde}, wherein signatures find utility in pricing problems~\cite{nonparam_pricing}.

Moreover, signatures have recently emerged as a subject of interest in the field of machine learning, particularly in the context of time series modeling. A comprehensive overview of these developments and surveys can be found in~\cite{lyons2023signature, primer_sig_ml}. Furthermore, researchers have endeavored to integrate signature computation into deep neural networks~\cite{deep_signature_transform}.  Additionally, novel machine learning approaches have been proposed for effectively modeling irregularly sampled time series~\cite{morrill2021generalised, morrill2021neural}, further broadening the scope and impact of signature-based techniques in this domain. Furthermore, the application of signatures as a tool in rough path theory for the study of fractional Brownian motions has been explored~\cite{tong2022learning}.

\paragraph{Transformers} Transformers~\cite{attention_is_all_you_need}, a recent advancement in deep neural network research, has gained widespread adoption in various domains, particularly in natural language processing~\cite{brown2020language} and computer vision~\cite{dosovitskiy2020image}. Additionally, there are several attempts to apply this approach to time-series data~\cite{li2020enhancing, zhou2021informer, wu2022autoformer,wen2023transformers} with a primary focus on addressing long-term prediction challenges. In contrast to the long-term prediction task emphasized in the aforementioned works, deep hedging employs an autoregressive approach to predict at each single step. It is worth noting that many of these transformer-based approaches have been employed in financial applications, leading to promising outcomes, as demonstrated by recent studies~\cite{barez2023exploring,arroyo2023deep,kisiel2022portfolio}. Nevertheless, it is important to highlight that certain variations of transformers have shown less than optimal performance when evaluated on financial datasets such as exchange indexes~\cite{Zeng2022AreTE}.

%% file: 3-model.tex
\begin{figure*}[!htbp]
    \includegraphics[width=0.9\textwidth]{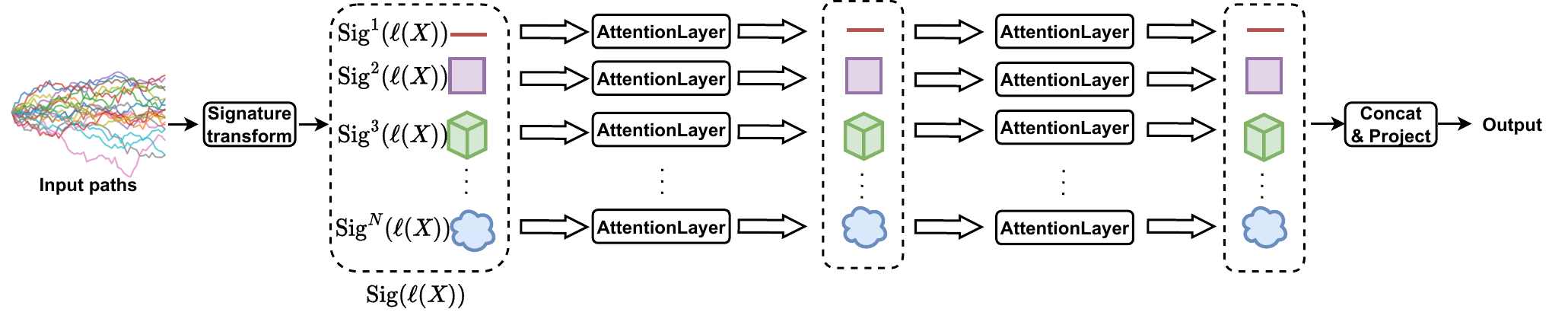}
    \caption{Overall architecture of \sigformer. Here we consider two layers of attentions. Signatures are truncated at $J$-th order. Given that the input paths are one-dimensional, for the purpose of illustration, we represent $\Sig^1(\ell(X)) \in \R$ as \includegraphics[scale=0.3]{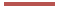}, $\Sig^2(\ell(X)) \in \R^{\otimes 2}$ as \includegraphics[scale=0.3]{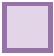}, $\Sig^3(\ell(X)) \in \R^{\otimes 3}$ as \includegraphics[scale=0.3]{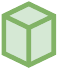} and so forth. Each layer layer in the plot, as in its original design \citep{attention_is_all_you_need}, contains three components: self-attention, multi-head attention, and feed-forward network.}
    \label{fig:main_fig}
\end{figure*}

\section{Signature Transformers}

This section presents our main proposed model, Signature Transformer or \sigformer.

\subsection{\sigformer}

This section focuses on the architecture of our main model, which we call SIGnature transFORMER or \sigformer.

\paragraph{Model specification}

Our goal is to construct a hedging strategy $\delta_k$ at time $t_k, k=0, \dots n$, which depends on the market information up to $t_{k-1}$, namely $I_{0}, \dots, I_{k-1}$. The primary aim of those hedging strategies is to minimize the risk as defined in Eq.~\eqref{eq:objective}. In our proposed model, we formulate it as a sequence-to-sequence modeling task. However, unlike recurrent approaches, we are interested in processing the entire input sequence $(I_0, \dots, I_{n-1})$ at once to produce the predicted sequence $(\delta_0, \dots, \delta_{n-1})$. We further denote $X_k$ in place of the market information $I_k$ in this section and write a sequence $X_0, \dots, X_k$ as $X_{0:k}$.


In essence, \sigformer is a transformer acting in the space of tensor algebra. We now describe its operations step by step. We first utilize the operator $\ell$ to  lift the input sequence $X_{0:n}$ while preserving the stream information~\cite{deep_signature_transform}. The operator $\ell$ is defined as:
\begin{equation*}
    \ell: X \mapsto (\ell^1(X), \ell^2(X), \dots, \ell^n(X)), \text{ where } \ell^k(X) \coloneqq X_{0:k}. 
\end{equation*}
Note that in practice, the sequence is padded with zeros in the beginning.

Next, by applying signature transformations to all the lifted sequences $\ell^k(X)$, for any $i=1, \dots, N$, we obtain the $i$-th level of the signature as
\begin{equation}
    \Sig^i(\ell(X)) \coloneqq \left(\Sig^i (\ell^1(X)), \dots, \Sig^i(\ell^n(X))\right) \in ((R^d)^{\otimes i})^{n}.
\end{equation}
The stream version of signatures can be considered as a sequence containing $n$ time steps in the space of $(\R^d)^{\otimes i}$.

Intuitively, the lift function $\ell$ preserves the stream information of sequence, because $\Sig(X_{0:n})$ is the summary up to time step $n$ but does not represent sequential structures.


At every individual signature level $\Sig^i(\ell(X))$, we apply 
\begin{equation}
\attentionlayer^i(\Sig^i(\ell(X))),
\end{equation}
where $\attentionlayer^i(\cdot)$ is a block containing all the components of the transformer described in~\S\ref{sec:attention}. Since $\Sig^i(\ell^k(X))$ stays in $(\R^d)^{\otimes i}$ for any time step $k$, we need to flatten it into $\R^{d ^ i}$ before applying projections to obtain queries, keys, values. To this end, we have
$$X \xrightarrow{\text{lift}} \ell(X) \xrightarrow{} (\Sig^i(\ell(X)))_{i=1}^N \xrightarrow{} (\attentionlayer^i(\Sig^i(\ell(X))))_{i=1}^N.$$
We apply the attention layer several times. In the final layer, we concatenate all transformed signatures and use a fully connected layer to get the output. Let us denote the whole architecture as $F^{\Sig}$. Figure~\ref{fig:main_fig} depicts an example of two-layer-attention \sigformer.

\begin{rem}
The primary motivation for designing separate attention layers for different signature levels is to equip \sigformer with flexibility in capturing the characteristic of sequence. That is, each signature level exhibits distinct geometric properties. For instance, the first level signature encodes changes over the interval, while the second level represents the L\'evy area, which corresponds to the area between the curve and the chord connecting its start and endpoints~\cite{morrill2021neural}.

We postulate that the geometric properties from the $i$-th order signature do not influence the decision of where to focus on the stream of another signature with different orders.

\end{rem}

\paragraph{Theoretical justification}It is worth noting that our construction of~\sigformer  possesses excellent approximation capabilities. \sigformer incorporates two main nonlinear transformations, namely $A \coloneqq \softmax(QK^\top / \sqrt{d_x})$ and a two-layer feed-forward network.
If we treat $A$ as a fixed matrix, we can conjecture the universal approximation capabilities of \sigformer. This is attributed to the universal approximation theorem for neural networks~\cite{pinkus_1999} and the universal approximation theorem for signatures~\cite[Theorem 3.4]{lyons2023signature} (see Proposition~\ref{prop:universal_nonlinearity}).

\subsection{Hedging with \sigformer}
Our hedge strategy uses $\delta^{\Sig} \coloneqq (\delta^{\Sig}_k)_{k=0, \dots, n-1}$ which relies on \sigformer to make decision.
Given market information $X$, formally, 
\begin{equation}
    \delta^{\Sig}_{0:n-1} = F^{\Sig}\left(X_{0:n-1})\right).
\end{equation}

Similar to~\cite{deep_hedging_rough_vol}, we train $F^{\Sig}$ by using the quadratic loss
\begin{equation*}
    \min_{\theta} \expect_{\mathbb{Q}} \left[ (p_0 + (\delta^{\Sig} \cdot S)_T - Z)^2\right],
\end{equation*}
where $p_0 = \expect_{Q}[Z]$. In our experiments, the underlying $\mathbb{Q}$ is a rough stochastic volatility model under European options.

The backbone neural network in~\citet{deep_hedging} w.r.t hedge strategy $\delta^{\rnn}$ can be designed in this form 
\begin{equation*}
    \delta^{\rnn}_k \coloneqq F_k(X_0, \dots, X_k, \delta^{\rnn}_{k-1}).
\end{equation*}
However, in the practical implementation,~\cite{deep_hedging} resorts to the \emph{semi-recurrence} neural network, a Markovian style, defined as $\delta^{\text{RNN}}_k \coloneqq F_k(X_k, \delta^{\text{RNN}}_{k-1})$. On the other hand, \cite{deep_hedging_rough_vol} presents an extension with hedge strategy $\delta_k = F_k(X_k, \delta_{k-1}, H_{k-1})$. The primary distinction between the semi-recurrent model~\cite{deep_hedging} and the recurrent model~\cite{deep_hedging_rough_vol} lies in the hidden state, $H_{k-1}$, which tries to capture the data's dynamics.

In contrast to using hidden states, our model makes a dynamic hedge at time $k$ using the signature of the entire trajectory up to $X_k$, denoted as  $\Sig(\ell^k(X))$ or $\Sig(X_{0:k})$.
As a result, our model can effectively handle data with memories like non-Markovian paths.



\begin{rem}
Compared to the recurrent architecture used in~\cite{deep_hedging_rough_vol}, transformer architectures or self-attention mechanisms process the entire sequence as a whole rather than recursively. This allows parallel computation and avoids the long dependency issues of the recurrent architecture. 
Transformers have proven to be more effective in modeling long sequences where RNNs tend to fall short of. The transformer's attention mechanism, for instance, in NLP tasks, demonstrates improved performance with longer input lengths, represented by a larger number of tokens~\cite{brown2020language}. 
However, it is important to note that the computational complexity of transformers is quadratic with respect to the length of the sequences.

The design of \sigformer is closely related to the work by~\citet{deep_signature_transform}, which enables a flexible combination of neural network components and signatures. In other words, \cite{deep_signature_transform} suggests that signature computation can be integrated as a part of deep neural networks. The design choices involve using multi-layer perceptrons (MLPs) or convolutional neural networks (CNNs) to either extract representations from signatures or to input signatures for computation.

In contrast, \sigformer follows a specific design, incorporating a sophisticated transformer approach that offers two key advantages. First, transformers stand out as an attractive model for sequential data compared to MLPs and CNNs. Second, while ~\cite{deep_signature_transform} treats all terms $\Sig^1(X), \dots, \Sig^N(X)$ equally through concatenation, \sigformer treats these terms individually, recognizing that they inherently represent different characteristics.

Note that we do not discuss the settings with constrained trading or transaction costs in this paper. However, it can be easy to extend our models for such cases as well.
    
\end{rem}

%% file: 4-experiment.tex
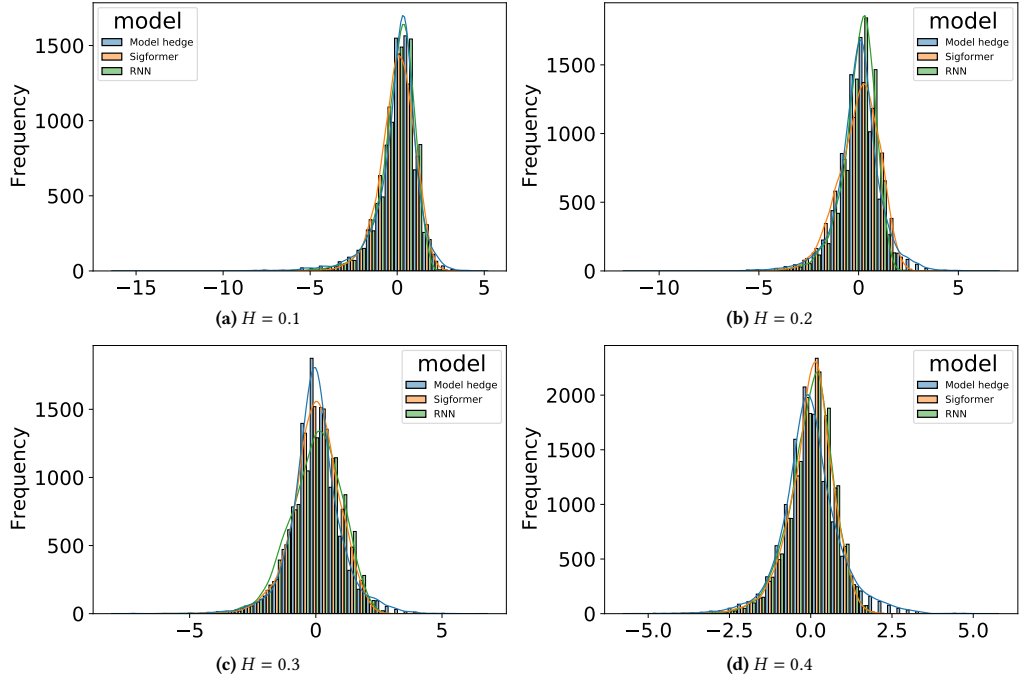
\begin{figure*}
    \centering
    \scalebox{0.8}{
    \begin{tikzpicture}
        \node[inner sep=0pt] (a) at (0,0) {\includegraphics[width=0.48\textwidth]{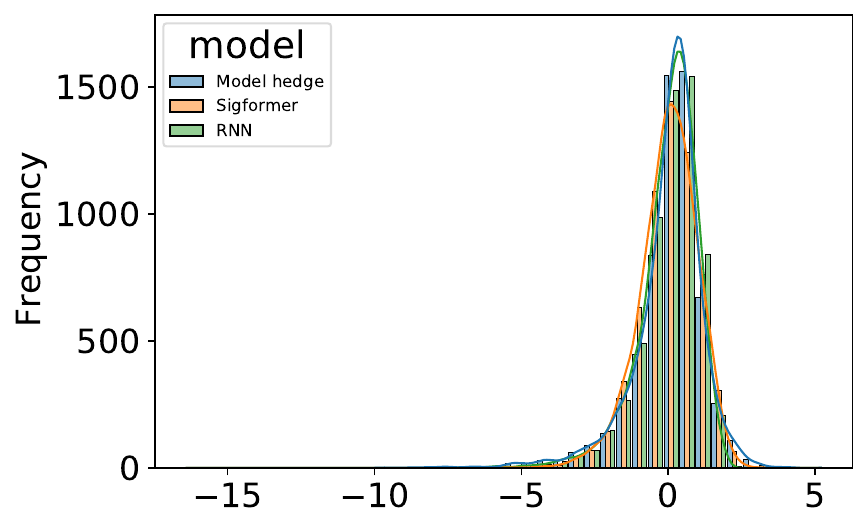}};

        \node[inner sep=0pt] (b) at (8.5,0) {\includegraphics[width=0.48\textwidth]{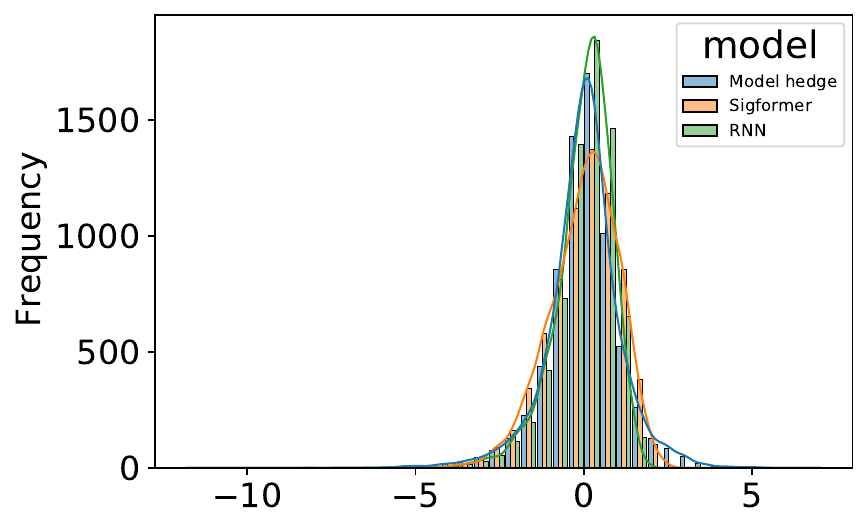}};

        \node[inner sep=0pt] (c) at (0, -5.7) {\includegraphics[width=0.48\textwidth]{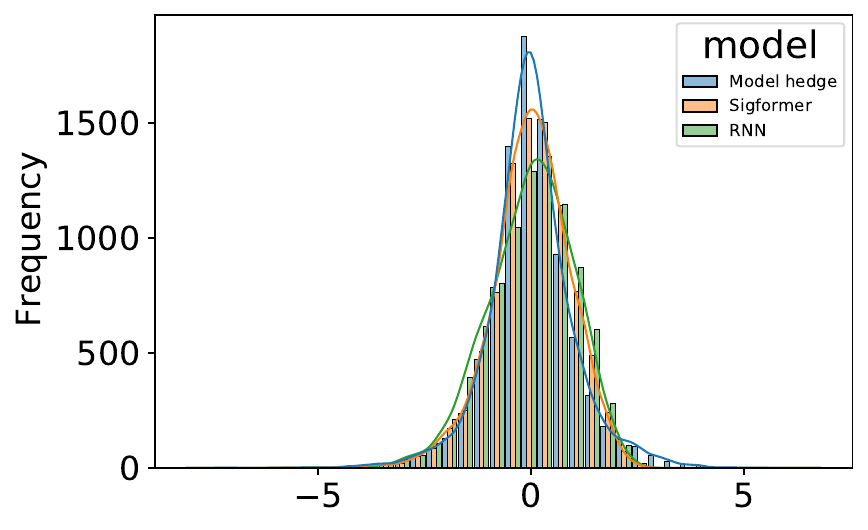}};

        \node[inner sep=0pt] (d) at (8.5, -5.7) {\includegraphics[width=0.48\textwidth]{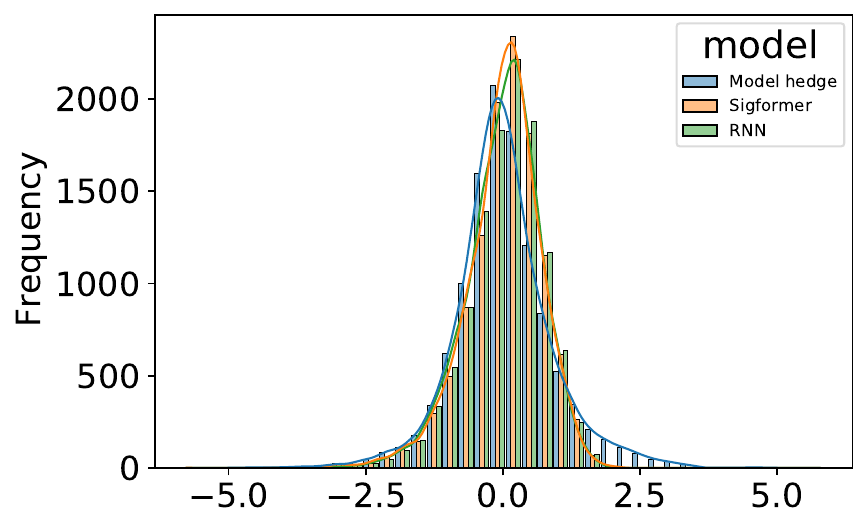}};

        \node (a_text) at (0, -2.8)    {\textbf{(a) $H=0.1$}};
        \node (b_text) at (8.5, -2.8)    {\textbf{(b) $H=0.2$}};
        \node (c_text) at (0, -8.55)    {\textbf{(c)} $H=0.3$};
        \node (d_text) at (8.5, -8.55)    {\textbf{(d)} $H=0.4$};
        
    \end{tikzpicture}
    }
    \caption{Comparison of risk-adjusted PnL between three models (enhanced clarity when zoomed in).}
    \label{fig:pnl}
\end{figure*}

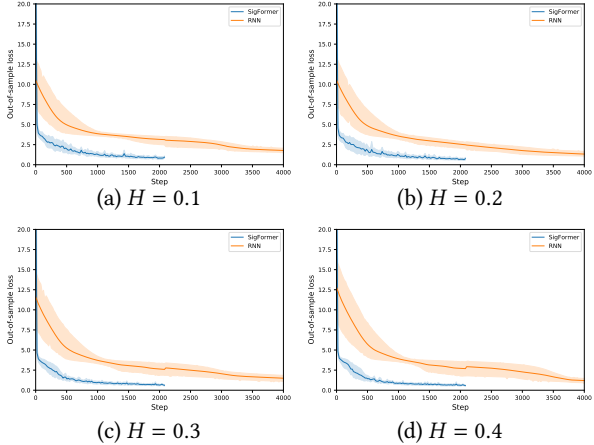
\begin{figure}
    \centering
    \begin{tikzpicture}
        \node[inner sep=0pt] (a) at (0,0) {\includegraphics[width=0.22\textwidth]{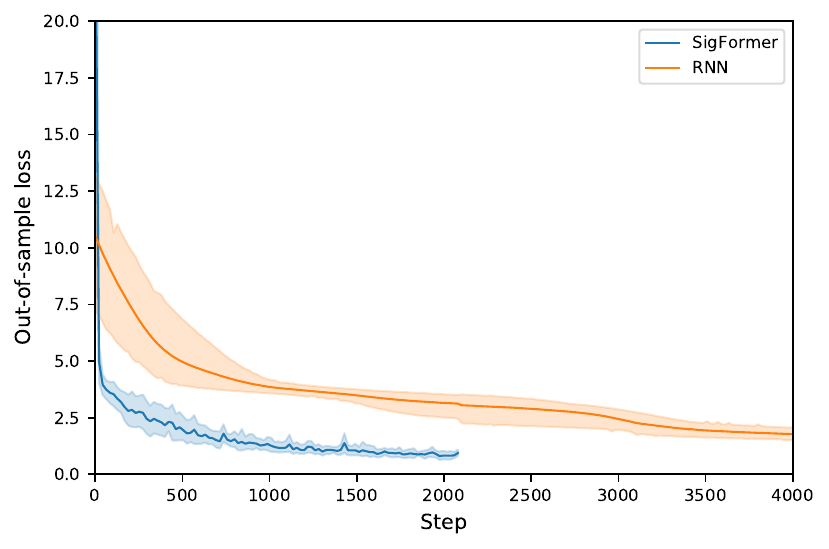}};

        \node[inner sep=0pt] (b) at (4, 0) {\includegraphics[width=0.22\textwidth]{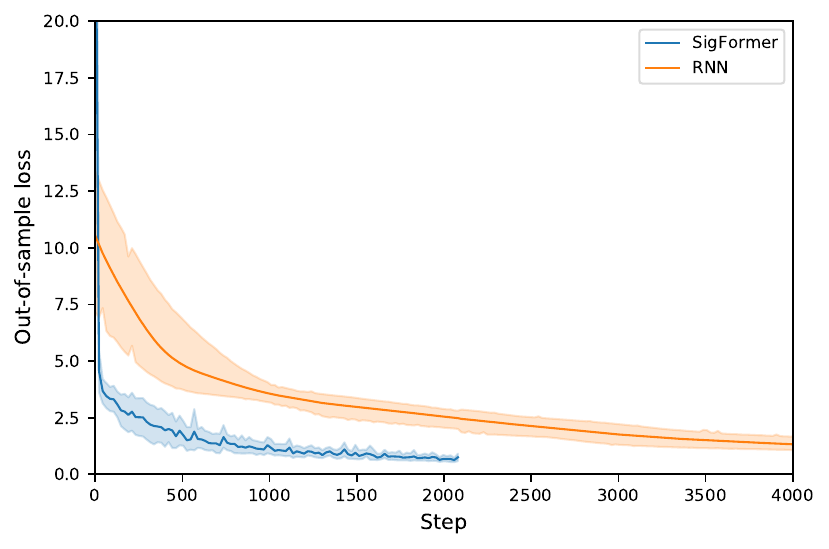}};

        \node (a_text) at (0, -1.4)    {(a) $H=0.1$};
        \node (b_text) at (4, -1.4)    {(b) $H=0.2$};

        \node[inner sep=0pt] (c) at (0,-3) {\includegraphics[width=0.22\textwidth]{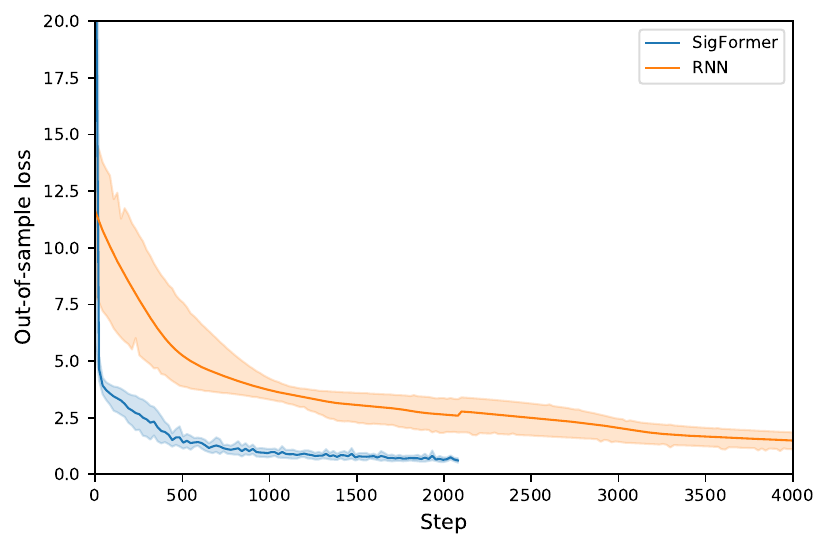}};

        \node[inner sep=0pt] (d) at (4, -3) {\includegraphics[width=0.22\textwidth]{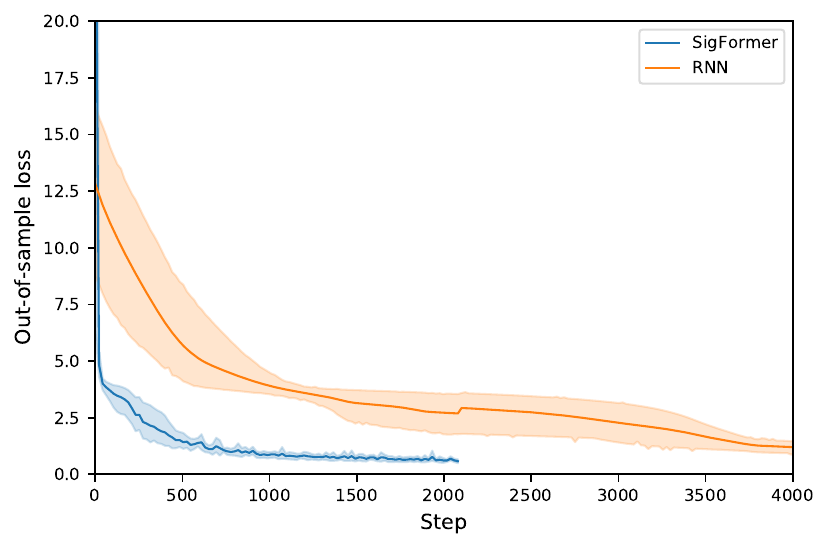}};

        \node (a_text) at (0, -4.5)    {(c) $H=0.3$};
        \node (b_text) at (4, -4.5)    {(d) $H=0.4$};
    \end{tikzpicture}
    \caption{Comparing out-of-sample loss in various Hurst parameter settings. We plot the loss curves with error bars computed over 5 independent runs. Clearly, \sigformer converges at a faster rate and does not require many training steps compared to the RNN approach.}
    \label{fig:convergence}
\end{figure}

\section{Experimental Results}
\label{sec:experiment}
This section presents empirical results comparing between the hedge strategy using \sigformer with the RNN approach from~\cite{deep_hedging_rough_vol}. Subsequently, we showcase a backtest conducted with real-world data for hedging S\&P500 index options.

Our source code is available at~\url{https://github.com/anh-tong/sigformer}, and it is implemented in JAX~\cite{jax2018github}. For computing signatures, we use \texttt{signax}~\footnote{https://pypi.org/project/signax/}. We observe that JAX offers faster running times in varies aspects such as simulating and solving stochastic differential equations, owing to its just-in-time (JIT) compilation features.

\subsection{Rough Bergomi model}
Consider a rough Bergomi model (rBergomi)~\cite{rbergomi_ref} as the underlying pricing model $\mathbb{Q}$. It is defined as
\begin{align}
    dS_t &= S_t\sqrt{V_t} (\sqrt{1 - \rho^2} dW_t + \rho dW_t^{\perp}),\\
    V_t &= \xi \exp\left(\eta W^H_t - \frac{1}{2}\eta^2 t^{2H}\right).
\end{align}
Here, $H$ is the Hurst parameter which indicates how irregular or ``rough'' the instantaneous variance process is, and $W, W^\perp$ are two independent Brownian motions. In brief, rBergomi is specified by four parameters: $H, \rho, \eta, \xi$. 
\paragraph{Perfect hedge} The portfolio of perfect hedge consists of stock price $S_t$ and its forward variance $\Theta_{T}^t = \sqrt{2H} \eta \int_0^T (s - r)^{H - \frac{1}{2}}dW_r$. According to~\cite{deep_hedging_rough_vol}, the contingent claim $Z_t = \expect(g(S_T)|\mathcal{F}_t)$ should be in a form $Z_t = u(t, S_{[0,t] },\Theta_{[t, T]}^t)$ with $\Theta_s^t = \sqrt{2H}\eta \int_0^t (s-r)^{H - \frac{1}{2}} dW_r$. The perfect hedge is written as
\begin{equation}
    dZ_t = \partial_x u(t, S_t, \Theta^t_{[t, T]}) dS_t + (T-t)^{1/2-H}\iprod{\partial_\omega u(t, S_t, \Theta^t_{[t, T]})}{a^t} d\Theta_T^t.
    \label{eq:perfect_hedging}
\end{equation} 
The second term in this equation is called path-wise Gateaux derivative and is the result of functional Itô formula~\cite{deep_hedging_rough_vol}.
Note that in our implementation, we do not use the finite-difference method to compute these derivatives like in~\cite{deep_hedging_rough_vol}. Instead, we leverage the capacities of auto-differentiation in JAX, as detailed in Appendix~\ref{appendix:compute_gradient}.

\paragraph{Simulation}Similar to~\cite{deep_hedging_rough_vol}, we adopt the approach proposed in~\cite{hybrid_scheme,turbo_charged} to generate samples for rBergomi models, represented as $S_t$. Additionally, we sample another instrument, known as the forward variance process $\Theta_{T_{\text{fwd}}}$, which has a longer maturity $T_{\text{fwd}}$. The sampling process for $\Theta_{T_{\text{fwd}}}$ follows $d\Theta^{t}_{T_{\text{fwd}}} = d\Theta^{t}_{T_{\text{fwd}}} \sqrt{2H}\eta (T_{\text{fwd}} - t)^{H - \frac{1}{2}}dW_t$, where $t \in [0, T]$.

\subsection{Empirical results of hedge strategy under rBergomi}
\label{sec:experiment_synthetic}
In the rough Bergomi model, we set $\rho=-0.7$, $\eta=-1.9$, and $\xi=0.235^2$, while varying the Hurst parameter as $H=0.1, 0.2, 0.3, 0.4$.

\paragraph{Data Generation}
In every training step, we generate $10^3$ new samples using \texttt{jax.random.fold\_in(key, current\_step)} in JAX code. This is considered as batch size for our training. For validation, we use fixed $10^4$ samples. And we use  $10^4$ out-of-sample for the test dataset.

\paragraph{Model Architecture}
For training, we use \sigformer with a truncated signature order of $3$. The attention mechanism consists of $12$ multi-heads and a total of $5$ attention layers. The recurrent neural network is constructed with $5$ hidden layers containing $128$ units each, using ReLU activation. All the model is trained with Adam method~\cite{adam_optim} with a learning rate $10^{-4}$.  We select the market information $I_k$ composed of two features: moneyness and volatility.

Figure~\ref{fig:pnl} depicts a comparison between the risk-adjusted profit and loss (PnL) of the model hedge, formulated using equation~\eqref{eq:perfect_hedging}, and two other approaches: the RNN approach presented in~\cite{deep_hedging_rough_vol} and our proposed approach - \sigformer. Notably, the upper tail of the PnL produced by \sigformer closely resembles a perfect hedge for the cases $H=0.1$ and $H=0.2$, when the price is known; however, it appears to be more irregular, showing tendencies of jumps. This can be attributed to the signature's ability to effectively model such properties. On the other hand, for $H=0.3$ and $H=0.4$, the results are comparable to those obtained with RNN models. Additionally, Figure~\ref{fig:convergence} illustrates that \sigformer exhibits faster convergence on validation datasets that RNNs.






\begin{table*}[!htpb]
    \centering
    \centering
    \scalebox{0.9}{
        \begin{tabular}{ccccccccccccc}
\multicolumn{13}{c}{} \\
& Jan & Feb & Mar & Apr & May & Jun & Jul & Aug & Sep & Oct & Nov & Dec \\
\hline
$H$                       & $0.071$& $0.072$& $0.051$ & $0.051$ & $0.064$ & $0.026$ & $0.069$ & $0.052$ & $0.092$& $0.025$ & $0.067$& $0.068$ \\
$\rho$                    & $-0.856$ & $-0.843$ & $-0.758$& $-0.837$ & $-0.851$ & $-0.749$ & $-0.841$ & $-0.806$& $-0.837$ & $-0.733$ & $-0.822$& $-0.835$ \\

$\eta$                    & $2.267$ & $2.284$ & $2.507$ & $2.235$ & $2.207$& $3.136$& $2.319$& $2.207$& $2.264$ & $1.953$& $1.976$& $2.195$\\
$\xi$                     & $0.050^2$& $0.050^2$& $0.383^2$ & $0.050^2$ & $0.271^2$ & $0.471^2$& $0.050^2$ & $0.172^2$ & $0.050^2$ & $0.068^2$ & $0.254^2$& $0.050^2$ \\
\hline

\end{tabular}
}
    
    \caption{Calibrated parameters of rough Bergomi model for the year of 2022.}
    \label{tab:rbergomi_parameter}
\end{table*}

\begin{figure}
    \centering
    \includegraphics[width=0.4\textwidth]{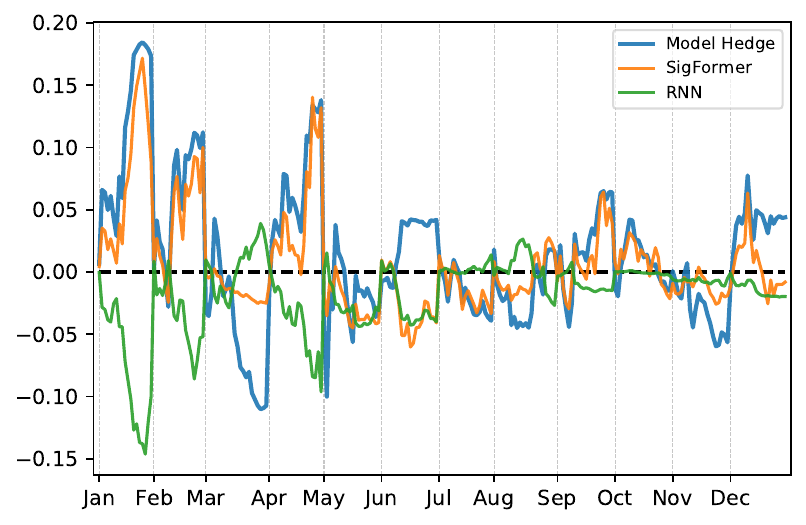}
    \caption{Wealth evolution.}
    \label{fig:pnl_evolution}
\end{figure}

\subsection{Backtest with real-world data}

This section presents an empirical result comparing our proposed model against the existing work including~\cite{deep_hedging_rough_vol}. We consider the rough Bergomi model for S\&P 500 index and are interested in hedging S\&P 500 with VIX index with maturity $1$ month.

We conduct a backtest using the data in the whole year of 2022. In detail, every month we calibrate the parameters of rough Bergomi models using~\cite{bayer2018deep}. Subsequently, we set the strike $K$ equal to the initial price $S_0$ of that month.

\paragraph{Data} We collect the market quotes\footnote{Downloadable from \url{www.optionsdx.com}} of S\&P 500 (SPX) in the year of 2022. We also gather VIX index of this year as the second instrument that proxies for the forward variance of rough Bergomi models.

\begin{figure}[!htpb]
    \centering
    \begin{tikzpicture}
        \node[inner sep=0pt] (a) at (0,0) {\includegraphics[width=0.23\textwidth]{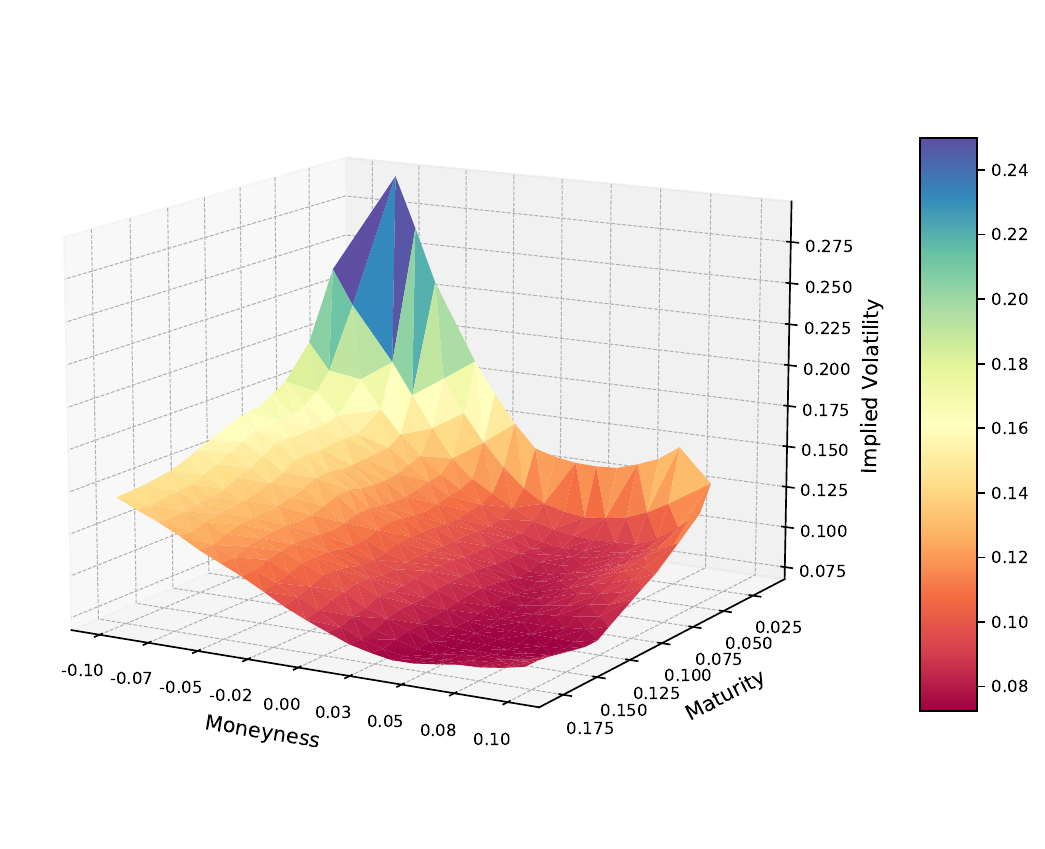}} ;

        \node[inner sep=0pt] (b) at (4, -0.2) {\includegraphics[width=0.23\textwidth]{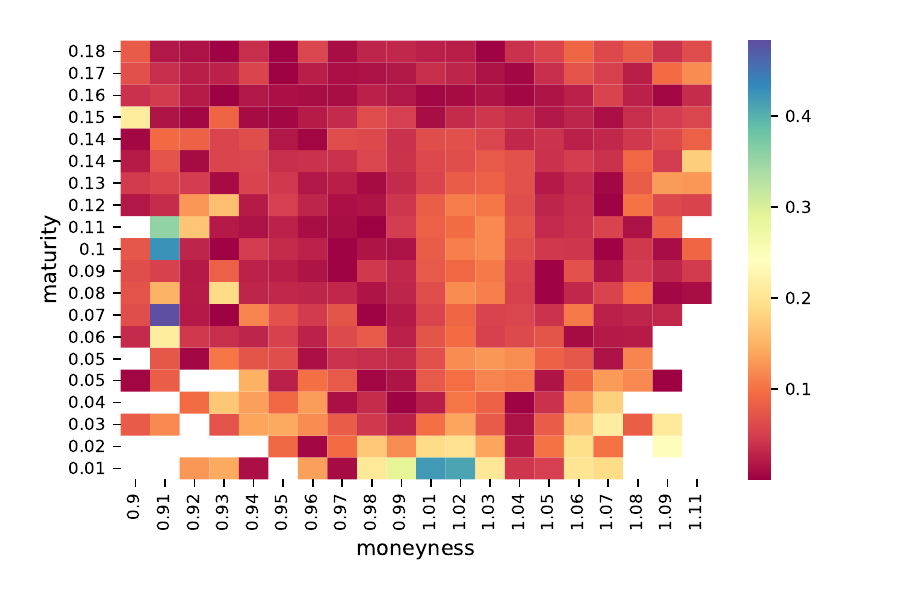}};

        \node (a_text) at (0, -1.7)    {(a)};
        \node (b_text) at (4, -1.7)    {(b)};
        
    \end{tikzpicture}
    
    \caption{Illustration of $\varphi_\NN$ in deep calibration. (a) Implied volatility surface produced by $\varphi_\NN$. (b) Relative error compared to the true implied volatility.}
    \label{fig:deep_calibration_implied_vol}
\end{figure}

\paragraph{rBergomi parameter calibration} We adopt the approach of~\citet{bayer2018deep} for calibrating the parameters $H, \eta, \rho$ and $\xi$. In this method, a neural network $\varphi_\NN$ is utilized to approximate implied volatility surfaces under rBergomi. After training $\varphi_{\NN}$ with $5 \times 10^5$ samples of rBergomi paths, we proceed with calibrating the model parameters using a Bayesian inference approach as described in~\cite[Section 5.2.1]{bayer2018deep}. The entire implementation is carried out in JAX for deep neural network training and Numpyro~\cite{numpyro} is employed for the Bayesian inference (see Table~\ref{tab:rbergomi_parameter} and Figure~\ref{fig:deep_calibration_implied_vol}).

Figure~\ref{fig:pnl_evolution} illustrates the wealth evolution of our hedge strategy based on \sigformer, the model hedge in equation~\eqref{eq:perfect_hedging}, alongside the results from the RNN approach~\cite{deep_hedging_rough_vol}. Remarkably, our model consistently generates positive PnL outcomes, in line with our observation regarding the upper tail of PnL distribution in~\S\ref{sec:experiment_synthetic} for small $H$.  Intriguingly, from January to April, our model and the RNN approach exhibit opposite PnL trends. We hypothesize that the hidden states learned by the RNN may not adequately encapsulate the information contained in the paths, while signatures are able to retain important characteristics of the path. During May and June, they yield similar results. Furthermore, in June, the performance of both the SigFormer and RNN models was inferior to that of the hedge model. We posit that this poorer performance stems from the unusually high magnitude of volatility ($\xi=0.471^2$, see Table~\ref{tab:rbergomi_parameter}) which poses difficulties for deep neural networks to process without additional preprocessing.

\subsection{Attention map}
\begin{figure}[!htpb]
    \centering
    \scalebox{0.8}{
    \begin{tikzpicture}
        \node (a) at (0,0) {\includegraphics[width=0.22\textwidth]{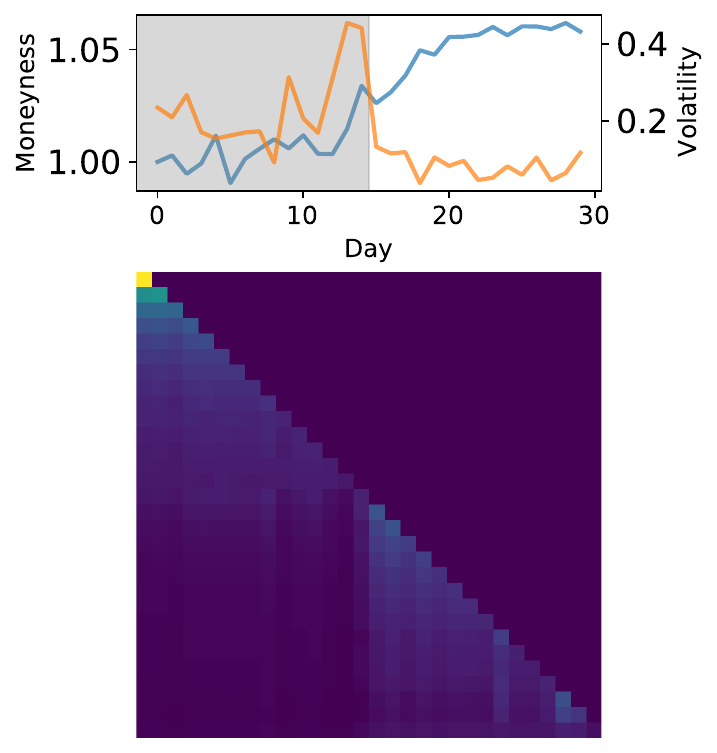}};



    \end{tikzpicture}
    }
    \caption{Visualize attention maps. The Attention map acts on $\Sig^3(X)$ (bottom) paring with the input as moneyness and volatility (top).}
    \label{fig:attenion_map}
\end{figure}

Next, we examine the attention maps in \sigformer.
$$A \coloneqq \softmax(QK^\top / \sqrt{d_x}).$$
These attention maps offers intriguing interpretability, revealing where the model allocates more attention while processing the input. Additionally, \sigformer allows us to differentiate attention maps between signature levels. In Figure~\ref{fig:attenion_map}, we present an example of an attention map generated for a given input. Remarkably, the attention map corresponding to $\Sig^3(X)$, exhibits distinct characteristics that are closely connected to the input.

\subsection{Ablation Study}

In this section, we conduct an ablation study to analyze the individual contributions of the signature computation and attention operator from the transformer in our proposed model. The study involves training models using each of these components separately to understand their impact on the overall performance.

First, we create a simplified signature model with a linear output given by:
\begin{equation*}
f_{\textrm{Signature}}(X) = \langle \Sig(X), W \rangle,
\end{equation*}
where $\Sig(X)$ represents the signature computation of the input data $X$, and $W \in T((\R^d))$ is the weight matrix.

\begin{figure}[!htpb]
    \centering
    \includegraphics[width=0.25\textwidth]{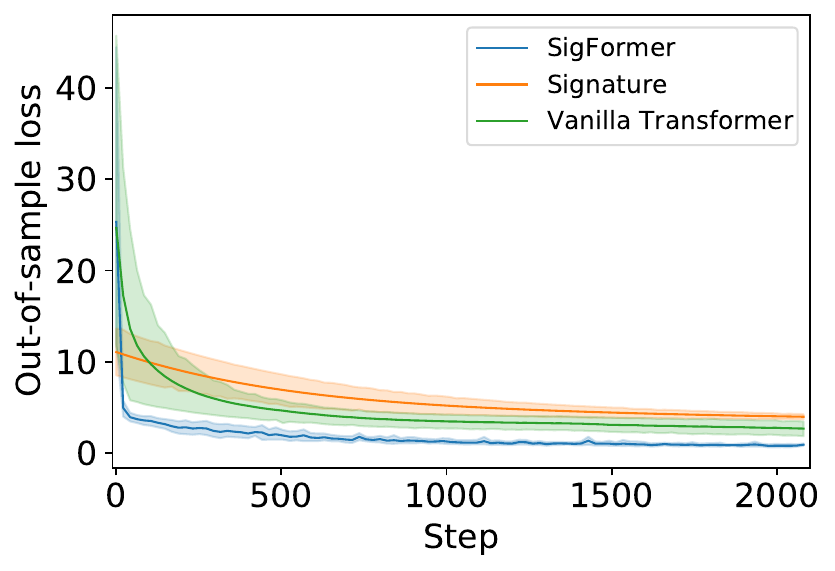}
    \caption{Out-of-sample loss among three models: \sigformer, vanilla transformer~\cite{attention_is_all_you_need}, and signature with linear. We consider $H=0.1$ in this experiment. The loss curves with error bars are computed over 5 independent runs. }
    \label{fig:ablation}
\end{figure}

For the transformer model, we utilize an encoder-style transformer with a fully connected layer at the last stage. And Figure~\ref{fig:ablation} provides clear evidence of our model outperforming the two baselines.

\begin{figure}[!htpb]
    \centering
    \begin{tikzpicture}
        \node[inner sep=0pt] (a) at (0,0) {\includegraphics[width=0.23\textwidth]{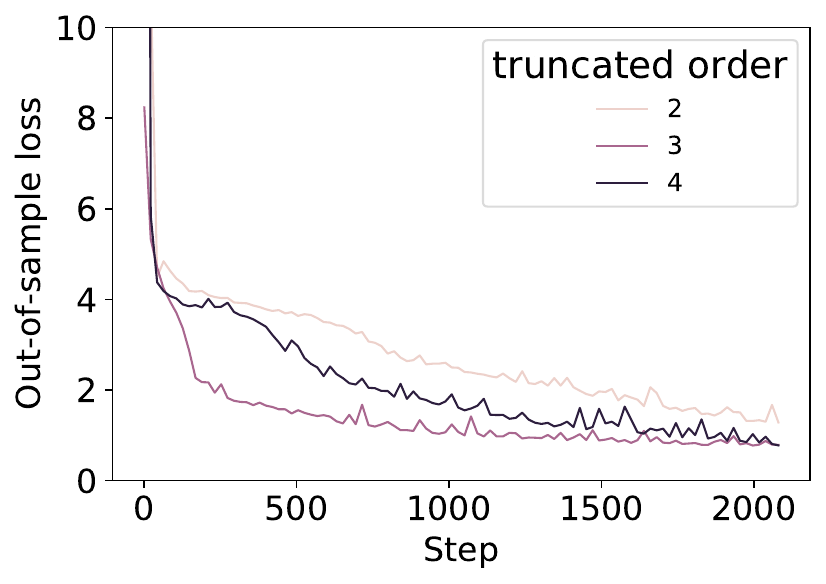}} ;

        \node[inner sep=0pt] (b) at (4, 0) {\includegraphics[width=0.23\textwidth]{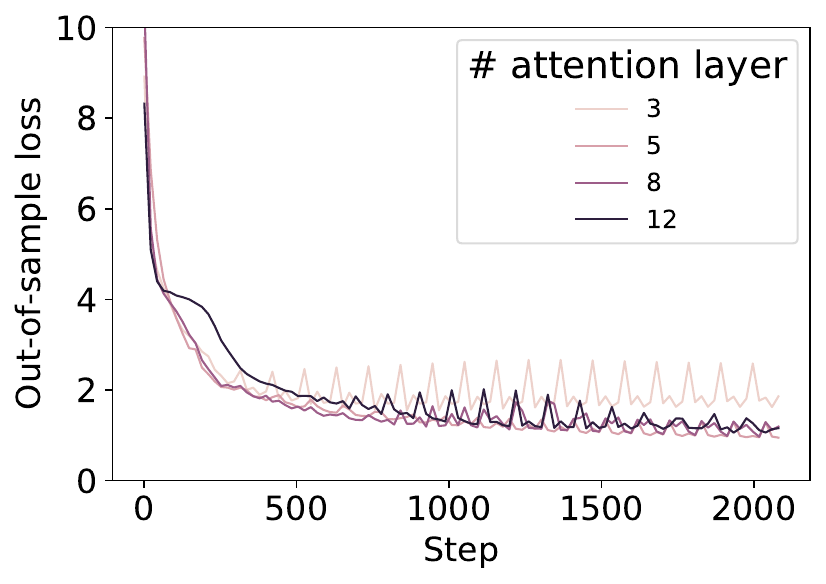}};

        \node (a_text) at (0, -1.6)    {(a)};
        \node (b_text) at (4, -1.6)    {(b)};
        
    \end{tikzpicture}

    \caption{Out-of-sample loss when (a) varying signature order and (b) varying the number of attention layers.}
    \label{fig:ablation_order_and_block}
\end{figure}

In a separate ablation experiment, we explored the impact of varying the number of truncated signature orders ($M$) and the number of attention blocks. Figure~\ref{fig:ablation_order_and_block} indicates that deeper models consistently outperform shallower ones. Additionally, the performance of signature order 3 is comparable to that of order 4, while the latter is significantly slower. Hence, we find signature order 3 to be a satisfactory choice for our model.


%% file: 5-conclusion.tex
\section{Conclusion and Discussion}
\paragraph{Conclusion.}
We presented \sigformer, a novel deep hedging model that is carefully built on the transformer architecture from machine learning and signature from rough path theory. As a result, we showed via our extensive experiments that \sigformer exhibits a strong advantage in handling irregularity, as compared to prior deep hedging models. We hope our research will draw more attention from both the finance community and machine learning community to the promising direction of exploring the advances in machine learning and rough path theory for addressing finance problems.

\paragraph{Limitations and Future Directions.} Our current model addresses deep hedging for a given portfolio and market state without adapting online to changes in our trading profiles and market conditions. That is, \sigformer is required to retrain on every new portfolio profile and market state. As our future revenue, we envision new adaptive models incorporating reinforcement learning (RL) for modeling online hedging strategies~\cite{buehler2023deep} with \sigformer (e.g., potentially via Decision Transformer~\citep{chen2021decision}). In particular, we can employ distributional RL~\citep{nguyen2021distributional} to estimate return distributions and thereby incorporate risk-adjusted returns conforming to human decision.




%% file: 6-appendix.tex
\section{Compute perfect delta hedge of rBergomi}
\label{appendix:compute_gradient}
Here is an example code computing the derivative {$\partial_x u(t, S_t, {\Theta}^t_{[t, T]})$} {and} Gautaex derivative {$\iprod{\partial_\omega u(t, S_t, {\Theta}^t_{[t, T]})}{a^t}$}. 

\begin{lstlisting}[language=Python, caption=Compute gradient]
import jax

def price_fn(S_t, epsilon):
    # make direction (T - t) ^ (H-1/2)
    a = ...
    Theta = Theta + a * epsilon
    # compute price given S and directional 
    ...

price_der, path_der = jax.grad(price_fn, (S_t, epsilon))

\end{lstlisting}
Note that we use the approximation in the above code
\begin{equation}
    \iprod{\partial_\omega u(t, S_t, {\Theta}^t_{[t, T]})}{a^t} \approx \left. \partial_\epsilon u(t, S_t, \{{\Theta}^t_i + \epsilon a_i^t\}_{i\in \mathcal{I}})\right|_{\epsilon=0}.
\end{equation}
Here, $\mathcal{I}$ is a discretization scheme over $[t,T]$. 

Comparing to finite-difference methods, using auto-differential framework is more accurate, and not restricted to approximation errors. However, it can be memory consuming and slower than finite-different counterparts.

%% file: main.bbl

\begin{thebibliography}{53}


\ifx \showCODEN    \undefined \def \showCODEN     #1{\unskip}     \fi
\ifx \showDOI      \undefined \def \showDOI       #1{#1}\fi
\ifx \showISBNx    \undefined \def \showISBNx     #1{\unskip}     \fi
\ifx \showISBNxiii \undefined \def \showISBNxiii  #1{\unskip}     \fi
\ifx \showISSN     \undefined \def \showISSN      #1{\unskip}     \fi
\ifx \showLCCN     \undefined \def \showLCCN      #1{\unskip}     \fi
\ifx \shownote     \undefined \def \shownote      #1{#1}          \fi
\ifx \showarticletitle \undefined \def \showarticletitle #1{#1}   \fi
\ifx \showURL      \undefined \def \showURL       {\relax}        \fi
\providecommand\bibfield[2]{#2}
\providecommand\bibinfo[2]{#2}
\providecommand\natexlab[1]{#1}
\providecommand\showeprint[2][]{arXiv:#2}

\bibitem[Arribas et~al\mbox{.}(2020)]%
        {sig_sde}
\bibfield{author}{\bibinfo{person}{Imanol~Perez Arribas},
  \bibinfo{person}{Cristopher Salvi}, {and} \bibinfo{person}{Lukasz Szpruch}.}
  \bibinfo{year}{2020}\natexlab{}.
\newblock \showarticletitle{Sig-SDEs Model for Quantitative Finance}. In
  \bibinfo{booktitle}{\emph{Proceedings of the First ACM International
  Conference on AI in Finance}} \emph{(\bibinfo{series}{ICAIF '20})}.
  \bibinfo{publisher}{Association for Computing Machinery},
  \bibinfo{address}{New York, NY, USA}, \bibinfo{numpages}{8}~pages.
\newblock


\bibitem[Arroyo et~al\mbox{.}(2023)]%
        {arroyo2023deep}
\bibfield{author}{\bibinfo{person}{Alvaro Arroyo}, \bibinfo{person}{Alvaro
  Cartea}, \bibinfo{person}{Fernando Moreno-Pino}, {and}
  \bibinfo{person}{Stefan Zohren}.} \bibinfo{year}{2023}\natexlab{}.
\newblock \bibinfo{title}{Deep Attentive Survival Analysis in Limit Order
  Books: Estimating Fill Probabilities with Convolutional-Transformers}.
\newblock
\newblock
\showeprint[arxiv]{2306.05479}~[q-fin.ST]


\bibitem[Barez et~al\mbox{.}(2023)]%
        {barez2023exploring}
\bibfield{author}{\bibinfo{person}{Fazl Barez}, \bibinfo{person}{Paul Bilokon},
  \bibinfo{person}{Arthur Gervais}, {and} \bibinfo{person}{Nikita Lisitsyn}.}
  \bibinfo{year}{2023}\natexlab{}.
\newblock \bibinfo{title}{Exploring the Advantages of Transformers for
  High-Frequency Trading}.
\newblock
\newblock
\showeprint[arxiv]{2302.13850}~[q-fin.ST]


\bibitem[Bayer et~al\mbox{.}(2016)]%
        {rbergomi_ref}
\bibfield{author}{\bibinfo{person}{Christian Bayer}, \bibinfo{person}{Peter
  Friz}, {and} \bibinfo{person}{Jim Gatheral}.}
  \bibinfo{year}{2016}\natexlab{}.
\newblock \showarticletitle{{Pricing under rough volatility}}.
\newblock \bibinfo{journal}{\emph{Quantitative Finance}} \bibinfo{volume}{16},
  \bibinfo{number}{6} (\bibinfo{date}{June} \bibinfo{year}{2016}),
  \bibinfo{pages}{887--904}.
\newblock


\bibitem[Bayer and Stemper(2018)]%
        {bayer2018deep}
\bibfield{author}{\bibinfo{person}{Christian Bayer} {and}
  \bibinfo{person}{Benjamin Stemper}.} \bibinfo{year}{2018}\natexlab{}.
\newblock \bibinfo{title}{Deep calibration of rough stochastic volatility
  models}.
\newblock
\newblock
\showeprint[arxiv]{1810.03399}~[q-fin.PR]


\bibitem[Bennedsen et~al\mbox{.}(2017)]%
        {hybrid_scheme}
\bibfield{author}{\bibinfo{person}{Mikkel Bennedsen}, \bibinfo{person}{Asger
  Lunde}, {and} \bibinfo{person}{Mikko~S. Pakkanen}.}
  \bibinfo{year}{2017}\natexlab{}.
\newblock \showarticletitle{Hybrid scheme for Brownian semistationary
  processes}.
\newblock \bibinfo{journal}{\emph{Finance and Stochastics}}
  \bibinfo{volume}{21}, \bibinfo{number}{4} (\bibinfo{date}{jun}
  \bibinfo{year}{2017}), \bibinfo{pages}{931--965}.
\newblock
\urldef\tempurl%
\url{https://doi.org/10.1007/s00780-017-0335-5}
\showDOI{\tempurl}


\bibitem[Bradbury et~al\mbox{.}(2018)]%
        {jax2018github}
\bibfield{author}{\bibinfo{person}{James Bradbury}, \bibinfo{person}{Roy
  Frostig}, \bibinfo{person}{Peter Hawkins}, \bibinfo{person}{Matthew~James
  Johnson}, \bibinfo{person}{Chris Leary}, \bibinfo{person}{Dougal Maclaurin},
  \bibinfo{person}{George Necula}, \bibinfo{person}{Adam Paszke},
  \bibinfo{person}{Jake Vander{P}las}, \bibinfo{person}{Skye
  Wanderman-{M}ilne}, {and} \bibinfo{person}{Qiao Zhang}.}
  \bibinfo{year}{2018}\natexlab{}.
\newblock \bibinfo{booktitle}{\emph{{JAX}: composable transformations of
  {P}ython+{N}um{P}y programs}}.
\newblock
\urldef\tempurl%
\url{http://github.com/google/jax}
\showURL{%
\tempurl}


\bibitem[Brown et~al\mbox{.}(2020)]%
        {brown2020language}
\bibfield{author}{\bibinfo{person}{Tom Brown}, \bibinfo{person}{Benjamin Mann},
  \bibinfo{person}{Nick Ryder}, \bibinfo{person}{Melanie Subbiah},
  \bibinfo{person}{Jared~D Kaplan}, \bibinfo{person}{Prafulla Dhariwal},
  \bibinfo{person}{Arvind Neelakantan}, \bibinfo{person}{Pranav Shyam},
  \bibinfo{person}{Girish Sastry}, \bibinfo{person}{Amanda Askell},
  {et~al\mbox{.}}} \bibinfo{year}{2020}\natexlab{}.
\newblock \showarticletitle{Language models are few-shot learners}.
\newblock \bibinfo{journal}{\emph{Advances in neural information processing
  systems}}  \bibinfo{volume}{33} (\bibinfo{year}{2020}),
  \bibinfo{pages}{1877--1901}.
\newblock


\bibitem[Buehler and Horvath(2022a)]%
        {buehler2022lecture_2}
\bibfield{author}{\bibinfo{person}{Hans Buehler} {and} \bibinfo{person}{Blanka
  Horvath}.} \bibinfo{year}{2022}\natexlab{a}.
\newblock \showarticletitle{Lecture Notes Learning to Trade I: Statistical
  Hedging}.
\newblock \bibinfo{journal}{\emph{Lecture Notes Learning to Trade I:
  Statistical Hedging (June 30, 2022)}} (\bibinfo{year}{2022}).
\newblock


\bibitem[Buehler and Horvath(2022b)]%
        {buehler2022lecture}
\bibfield{author}{\bibinfo{person}{Hans Buehler} {and} \bibinfo{person}{Blanka
  Horvath}.} \bibinfo{year}{2022}\natexlab{b}.
\newblock \showarticletitle{Lecture Notes Learning to Trade II: Deep Hedging}.
\newblock \bibinfo{journal}{\emph{Lecture Notes Learning to Trade II: Deep
  Hedging (June 30, 2022)}} (\bibinfo{year}{2022}).
\newblock


\bibitem[Bühler et~al\mbox{.}(2018)]%
        {deep_hedging}
\bibfield{author}{\bibinfo{person}{Hans Bühler}, \bibinfo{person}{Lukas
  Gonon}, \bibinfo{person}{Josef Teichmann}, {and} \bibinfo{person}{Ben Wood}.}
  \bibinfo{year}{2018}\natexlab{}.
\newblock \bibinfo{title}{Deep Hedging}.
\newblock
\newblock


\bibitem[Bühler et~al\mbox{.}(2022)]%
        {buehler2022deep}
\bibfield{author}{\bibinfo{person}{Hans Bühler}, \bibinfo{person}{Phillip
  Murray}, \bibinfo{person}{Mikko~S. Pakkanen}, {and} \bibinfo{person}{Ben
  Wood}.} \bibinfo{year}{2022}\natexlab{}.
\newblock \bibinfo{title}{Deep Hedging: Learning to Remove the Drift under
  Trading Frictions with Minimal Equivalent Near-Martingale Measures}.
\newblock
\newblock
\showeprint[arxiv]{2111.07844}~[q-fin.CP]


\bibitem[Bühler et~al\mbox{.}(2023)]%
        {buehler2023deep}
\bibfield{author}{\bibinfo{person}{Hans Bühler}, \bibinfo{person}{Phillip
  Murray}, {and} \bibinfo{person}{Ben Wood}.} \bibinfo{year}{2023}\natexlab{}.
\newblock \bibinfo{title}{Deep Bellman Hedging}.
\newblock
\newblock
\showeprint[arxiv]{2207.00932}~[q-fin.CP]


\bibitem[Chen et~al\mbox{.}(2021)]%
        {chen2021decision}
\bibfield{author}{\bibinfo{person}{Lili Chen}, \bibinfo{person}{Kevin Lu},
  \bibinfo{person}{Aravind Rajeswaran}, \bibinfo{person}{Kimin Lee},
  \bibinfo{person}{Aditya Grover}, \bibinfo{person}{Misha Laskin},
  \bibinfo{person}{Pieter Abbeel}, \bibinfo{person}{Aravind Srinivas}, {and}
  \bibinfo{person}{Igor Mordatch}.} \bibinfo{year}{2021}\natexlab{}.
\newblock \showarticletitle{Decision transformer: Reinforcement learning via
  sequence modeling}.
\newblock \bibinfo{journal}{\emph{Advances in neural information processing
  systems}}  \bibinfo{volume}{34} (\bibinfo{year}{2021}),
  \bibinfo{pages}{15084--15097}.
\newblock


\bibitem[Chevyrev and Kormilitzin(2016)]%
        {primer_sig_ml}
\bibfield{author}{\bibinfo{person}{Ilya Chevyrev} {and} \bibinfo{person}{Andrey
  Kormilitzin}.} \bibinfo{year}{2016}\natexlab{}.
\newblock \bibinfo{title}{A Primer on the Signature Method in Machine
  Learning}.
\newblock
\newblock


\bibitem[Dosovitskiy et~al\mbox{.}(2020)]%
        {dosovitskiy2020image}
\bibfield{author}{\bibinfo{person}{Alexey Dosovitskiy}, \bibinfo{person}{Lucas
  Beyer}, \bibinfo{person}{Alexander Kolesnikov}, \bibinfo{person}{Dirk
  Weissenborn}, \bibinfo{person}{Xiaohua Zhai}, \bibinfo{person}{Thomas
  Unterthiner}, \bibinfo{person}{Mostafa Dehghani}, \bibinfo{person}{Matthias
  Minderer}, \bibinfo{person}{Georg Heigold}, \bibinfo{person}{Sylvain Gelly},
  {et~al\mbox{.}}} \bibinfo{year}{2020}\natexlab{}.
\newblock \showarticletitle{An image is worth 16x16 words: Transformers for
  image recognition at scale}.
\newblock \bibinfo{journal}{\emph{arXiv preprint arXiv:2010.11929}}
  (\bibinfo{year}{2020}).
\newblock


\bibitem[Friz and Hairer(2020)]%
        {friz2020course}
\bibfield{author}{\bibinfo{person}{Peter~K Friz} {and} \bibinfo{person}{Martin
  Hairer}.} \bibinfo{year}{2020}\natexlab{}.
\newblock \bibinfo{booktitle}{\emph{A Course on Rough Paths: With an
  Introduction to Regularity Structures}}.
\newblock \bibinfo{publisher}{Springer Nature}.
\newblock


\bibitem[Friz and Victoir(2010)]%
        {friz2010multidimensional}
\bibfield{author}{\bibinfo{person}{Peter~K Friz} {and}
  \bibinfo{person}{Nicolas~B Victoir}.} \bibinfo{year}{2010}\natexlab{}.
\newblock \bibinfo{booktitle}{\emph{Multidimensional stochastic processes as
  rough paths: theory and applications}}. Vol.~\bibinfo{volume}{120}.
\newblock \bibinfo{publisher}{Cambridge University Press}.
\newblock


\bibitem[Gatheral et~al\mbox{.}(2014)]%
        {gatheral2014volatility}
\bibfield{author}{\bibinfo{person}{Jim Gatheral}, \bibinfo{person}{Thibault
  Jaisson}, {and} \bibinfo{person}{Mathieu Rosenbaum}.}
  \bibinfo{year}{2014}\natexlab{}.
\newblock \bibinfo{title}{Volatility is rough}.
\newblock
\newblock
\showeprint[arxiv]{1410.3394}~[q-fin.ST]


\bibitem[Geva et~al\mbox{.}(2021)]%
        {geva2021transformer}
\bibfield{author}{\bibinfo{person}{Mor Geva}, \bibinfo{person}{Roei Schuster},
  \bibinfo{person}{Jonathan Berant}, {and} \bibinfo{person}{Omer Levy}.}
  \bibinfo{year}{2021}\natexlab{}.
\newblock \bibinfo{title}{Transformer Feed-Forward Layers Are Key-Value
  Memories}.
\newblock
\newblock
\showeprint[arxiv]{2012.14913}~[cs.CL]


\bibitem[Heston(1993)]%
        {heston1993closed}
\bibfield{author}{\bibinfo{person}{Steven~L Heston}.}
  \bibinfo{year}{1993}\natexlab{}.
\newblock \showarticletitle{A closed-form solution for options with stochastic
  volatility with applications to bond and currency options}.
\newblock \bibinfo{journal}{\emph{The review of financial studies}}
  \bibinfo{volume}{6}, \bibinfo{number}{2} (\bibinfo{year}{1993}),
  \bibinfo{pages}{327--343}.
\newblock


\bibitem[Hornik et~al\mbox{.}(1989)]%
        {hornik1989multilayer}
\bibfield{author}{\bibinfo{person}{Kurt Hornik}, \bibinfo{person}{Maxwell
  Stinchcombe}, {and} \bibinfo{person}{Halbert White}.}
  \bibinfo{year}{1989}\natexlab{}.
\newblock \showarticletitle{Multilayer feedforward networks are universal
  approximators}.
\newblock \bibinfo{journal}{\emph{Neural networks}} \bibinfo{volume}{2},
  \bibinfo{number}{5} (\bibinfo{year}{1989}), \bibinfo{pages}{359--366}.
\newblock


\bibitem[Horvath et~al\mbox{.}(2021)]%
        {deep_hedging_rough_vol}
\bibfield{author}{\bibinfo{person}{Blanka Horvath}, \bibinfo{person}{Josef
  Teichmann}, {and} \bibinfo{person}{Zan Zuric}.}
  \bibinfo{year}{2021}\natexlab{}.
\newblock \bibinfo{title}{Deep Hedging under Rough Volatility}.
\newblock
\newblock


\bibitem[Ilhan et~al\mbox{.}(2009)]%
        {ilhan2009optimal}
\bibfield{author}{\bibinfo{person}{Ayta{\c{c}} Ilhan}, \bibinfo{person}{Mattias
  Jonsson}, {and} \bibinfo{person}{Ronnie Sircar}.}
  \bibinfo{year}{2009}\natexlab{}.
\newblock \showarticletitle{Optimal static-dynamic hedges for exotic options
  under convex risk measures}.
\newblock \bibinfo{journal}{\emph{Stochastic Processes and their Applications}}
  \bibinfo{volume}{119}, \bibinfo{number}{10} (\bibinfo{year}{2009}),
  \bibinfo{pages}{3608--3632}.
\newblock


\bibitem[Kidger et~al\mbox{.}(2019)]%
        {deep_signature_transform}
\bibfield{author}{\bibinfo{person}{Patrick Kidger}, \bibinfo{person}{Patric
  Bonnier}, \bibinfo{person}{Imanol Perez~Arribas}, \bibinfo{person}{Cristopher
  Salvi}, {and} \bibinfo{person}{Terry Lyons}.}
  \bibinfo{year}{2019}\natexlab{}.
\newblock \showarticletitle{Deep Signature Transforms}. In
  \bibinfo{booktitle}{\emph{Advances in Neural Information Processing
  Systems}}, \bibfield{editor}{\bibinfo{person}{H.~Wallach},
  \bibinfo{person}{H.~Larochelle}, \bibinfo{person}{A.~Beygelzimer},
  \bibinfo{person}{F.~d\textquotesingle Alch\'{e}-Buc},
  \bibinfo{person}{E.~Fox}, {and} \bibinfo{person}{R.~Garnett}} (Eds.).
  \bibinfo{pages}{3099--3109}.
\newblock


\bibitem[Kidger and Lyons(2021)]%
        {kidger2021signatory}
\bibfield{author}{\bibinfo{person}{Patrick Kidger} {and} \bibinfo{person}{Terry
  Lyons}.} \bibinfo{year}{2021}\natexlab{}.
\newblock \showarticletitle{{S}ignatory: differentiable computations of the
  signature and logsignature transforms, on both {CPU} and {GPU}}. In
  \bibinfo{booktitle}{\emph{International Conference on Learning
  Representations}}.
\newblock


\bibitem[Kingma and Ba(2014)]%
        {adam_optim}
\bibfield{author}{\bibinfo{person}{Diederik~P. Kingma} {and}
  \bibinfo{person}{Jimmy Ba}.} \bibinfo{year}{2014}\natexlab{}.
\newblock \bibinfo{title}{Adam: A Method for Stochastic Optimization}.
\newblock
\newblock
\urldef\tempurl%
\url{https://doi.org/10.48550/ARXIV.1412.6980}
\showDOI{\tempurl}


\bibitem[Kiraly and Oberhauser(2019)]%
        {signature_kernel_1}
\bibfield{author}{\bibinfo{person}{Franz~J. Kiraly} {and}
  \bibinfo{person}{Harald Oberhauser}.} \bibinfo{year}{2019}\natexlab{}.
\newblock \showarticletitle{Kernels for Sequentially Ordered Data}.
\newblock \bibinfo{journal}{\emph{Journal of Machine Learning Research}}
  \bibinfo{volume}{20}, \bibinfo{number}{31} (\bibinfo{year}{2019}),
  \bibinfo{pages}{1--45}.
\newblock


\bibitem[Kisiel and Gorse(2022)]%
        {kisiel2022portfolio}
\bibfield{author}{\bibinfo{person}{Damian Kisiel} {and} \bibinfo{person}{Denise
  Gorse}.} \bibinfo{year}{2022}\natexlab{}.
\newblock \bibinfo{title}{Portfolio Transformer for Attention-Based Asset
  Allocation}.
\newblock
\newblock
\showeprint[arxiv]{2206.03246}~[q-fin.PM]


\bibitem[Li et~al\mbox{.}(2020)]%
        {li2020enhancing}
\bibfield{author}{\bibinfo{person}{Shiyang Li}, \bibinfo{person}{Xiaoyong Jin},
  \bibinfo{person}{Yao Xuan}, \bibinfo{person}{Xiyou Zhou},
  \bibinfo{person}{Wenhu Chen}, \bibinfo{person}{Yu-Xiang Wang}, {and}
  \bibinfo{person}{Xifeng Yan}.} \bibinfo{year}{2020}\natexlab{}.
\newblock \bibinfo{title}{Enhancing the Locality and Breaking the Memory
  Bottleneck of Transformer on Time Series Forecasting}.
\newblock
\newblock
\showeprint[arxiv]{1907.00235}~[cs.LG]


\bibitem[Limmer and Horvath(2023)]%
        {robust_hedging_GANs}
\bibfield{author}{\bibinfo{person}{Yannick Limmer} {and}
  \bibinfo{person}{Blanka Horvath}.} \bibinfo{year}{2023}\natexlab{}.
\newblock \bibinfo{title}{Robust Hedging GANs}.
\newblock
\newblock
\showeprint[arxiv]{2307.02310}~[q-fin.CP]


\bibitem[Lyons(2014)]%
        {Lyons2014RoughPS}
\bibfield{author}{\bibinfo{person}{Terry Lyons}.}
  \bibinfo{year}{2014}\natexlab{}.
\newblock \showarticletitle{Rough paths, Signatures and the modelling of
  functions on streams}.
\newblock \bibinfo{journal}{\emph{arXiv: Probability}} (\bibinfo{year}{2014}).
\newblock


\bibitem[Lyons and McLeod(2023)]%
        {lyons2023signature}
\bibfield{author}{\bibinfo{person}{Terry Lyons} {and}
  \bibinfo{person}{Andrew~D. McLeod}.} \bibinfo{year}{2023}\natexlab{}.
\newblock \bibinfo{title}{Signature Methods in Machine Learning}.
\newblock
\newblock
\showeprint[arxiv]{2206.14674}~[stat.ML]


\bibitem[Lyons et~al\mbox{.}(2019)]%
        {nonparam_pricing}
\bibfield{author}{\bibinfo{person}{Terry Lyons}, \bibinfo{person}{Sina Nejad},
  {and} \bibinfo{person}{Imanol~Perez Arribas}.}
  \bibinfo{year}{2019}\natexlab{}.
\newblock \bibinfo{title}{Nonparametric pricing and hedging of exotic
  derivatives}.
\newblock
\newblock


\bibitem[Lyons(1998)]%
        {Lyons1998}
\bibfield{author}{\bibinfo{person}{Terry~J. Lyons}.}
  \bibinfo{year}{1998}\natexlab{}.
\newblock \showarticletitle{Differential equations driven by rough signals.}
\newblock \bibinfo{journal}{\emph{Revista Matemática Iberoamericana}}
  \bibinfo{volume}{14}, \bibinfo{number}{2} (\bibinfo{year}{1998}),
  \bibinfo{pages}{215--310}.
\newblock


\bibitem[Mandelbrot and Van~Ness(1968)]%
        {mandelbrot1968fractional}
\bibfield{author}{\bibinfo{person}{Benoit~B Mandelbrot} {and}
  \bibinfo{person}{John~W Van~Ness}.} \bibinfo{year}{1968}\natexlab{}.
\newblock \showarticletitle{Fractional Brownian motions, fractional noises and
  applications}.
\newblock \bibinfo{journal}{\emph{SIAM review}} \bibinfo{volume}{10},
  \bibinfo{number}{4} (\bibinfo{year}{1968}), \bibinfo{pages}{422--437}.
\newblock


\bibitem[McCrickerd and Pakkanen(2018)]%
        {turbo_charged}
\bibfield{author}{\bibinfo{person}{Ryan McCrickerd} {and}
  \bibinfo{person}{Mikko~S. Pakkanen}.} \bibinfo{year}{2018}\natexlab{}.
\newblock \showarticletitle{Turbocharging Monte Carlo pricing for the rough
  Bergomi model}.
\newblock \bibinfo{journal}{\emph{Quantitative Finance}} \bibinfo{volume}{18},
  \bibinfo{number}{11} (\bibinfo{date}{apr} \bibinfo{year}{2018}),
  \bibinfo{pages}{1877--1886}.
\newblock
\urldef\tempurl%
\url{https://doi.org/10.1080/14697688.2018.1459812}
\showDOI{\tempurl}


\bibitem[Morrill et~al\mbox{.}(2021a)]%
        {morrill2021generalised}
\bibfield{author}{\bibinfo{person}{James Morrill}, \bibinfo{person}{Adeline
  Fermanian}, \bibinfo{person}{Patrick Kidger}, {and} \bibinfo{person}{Terry
  Lyons}.} \bibinfo{year}{2021}\natexlab{a}.
\newblock \bibinfo{title}{A Generalised Signature Method for Multivariate Time
  Series Feature Extraction}.
\newblock
\newblock
\showeprint[arxiv]{2006.00873}~[cs.LG]


\bibitem[Morrill et~al\mbox{.}(2021b)]%
        {morrill2021neural}
\bibfield{author}{\bibinfo{person}{James Morrill}, \bibinfo{person}{Cristopher
  Salvi}, \bibinfo{person}{Patrick Kidger}, \bibinfo{person}{James Foster},
  {and} \bibinfo{person}{Terry Lyons}.} \bibinfo{year}{2021}\natexlab{b}.
\newblock \bibinfo{title}{Neural Rough Differential Equations for Long Time
  Series}.
\newblock
\newblock
\showeprint[arxiv]{2009.08295}~[cs.LG]


\bibitem[Nguyen-Tang et~al\mbox{.}(2021)]%
        {nguyen2021distributional}
\bibfield{author}{\bibinfo{person}{Thanh Nguyen-Tang}, \bibinfo{person}{Sunil
  Gupta}, {and} \bibinfo{person}{Svetha Venkatesh}.}
  \bibinfo{year}{2021}\natexlab{}.
\newblock \showarticletitle{Distributional reinforcement learning via moment
  matching}. In \bibinfo{booktitle}{\emph{Proceedings of the AAAI Conference on
  Artificial Intelligence}}, Vol.~\bibinfo{volume}{35}.
  \bibinfo{pages}{9144--9152}.
\newblock


\bibitem[Phan et~al\mbox{.}(2019)]%
        {numpyro}
\bibfield{author}{\bibinfo{person}{Du Phan}, \bibinfo{person}{Neeraj Pradhan},
  {and} \bibinfo{person}{Martin Jankowiak}.} \bibinfo{year}{2019}\natexlab{}.
\newblock \showarticletitle{Composable Effects for Flexible and Accelerated
  Probabilistic Programming in NumPyro}.
\newblock \bibinfo{journal}{\emph{arXiv preprint arXiv:1912.11554}}
  (\bibinfo{year}{2019}).
\newblock


\bibitem[Pinkus(1999)]%
        {pinkus_1999}
\bibfield{author}{\bibinfo{person}{Allan Pinkus}.}
  \bibinfo{year}{1999}\natexlab{}.
\newblock \showarticletitle{Approximation theory of the MLP model in neural
  networks}.
\newblock \bibinfo{journal}{\emph{Acta Numerica}}  \bibinfo{volume}{8}
  (\bibinfo{year}{1999}), \bibinfo{pages}{143–195}.
\newblock
\urldef\tempurl%
\url{https://doi.org/10.1017/S0962492900002919}
\showDOI{\tempurl}


\bibitem[Reizenstein and Graham(2020)]%
        {iisignature}
\bibfield{author}{\bibinfo{person}{Jeremy~F. Reizenstein} {and}
  \bibinfo{person}{Benjamin Graham}.} \bibinfo{year}{2020}\natexlab{}.
\newblock \showarticletitle{Algorithm 1004: The Iisignature Library: Efficient
  Calculation of Iterated-Integral Signatures and Log Signatures}.
\newblock \bibinfo{journal}{\emph{ACM Trans. Math. Softw.}}
  \bibinfo{volume}{46}, \bibinfo{number}{1} (\bibinfo{year}{2020}).
\newblock
\showISSN{0098-3500}


\bibitem[Salvi et~al\mbox{.}(2021)]%
        {signature_kernel_2}
\bibfield{author}{\bibinfo{person}{Cristopher Salvi}, \bibinfo{person}{Thomas
  Cass}, \bibinfo{person}{James Foster}, \bibinfo{person}{Terry Lyons}, {and}
  \bibinfo{person}{Weixin Yang}.} \bibinfo{year}{2021}\natexlab{}.
\newblock \showarticletitle{The Signature Kernel Is the Solution of a Goursat
  {PDE}}.
\newblock \bibinfo{journal}{\emph{{SIAM} Journal on Mathematics of Data
  Science}} (\bibinfo{date}{jan} \bibinfo{year}{2021}),
  \bibinfo{pages}{873--899}.
\newblock


\bibitem[Tong et~al\mbox{.}(2022)]%
        {tong2022learning}
\bibfield{author}{\bibinfo{person}{Anh Tong}, \bibinfo{person}{Thanh
  Nguyen-Tang}, \bibinfo{person}{Toan Tran}, {and} \bibinfo{person}{Jaesik
  Choi}.} \bibinfo{year}{2022}\natexlab{}.
\newblock \showarticletitle{Learning Fractional White Noises in Neural
  Stochastic Differential Equations}. In \bibinfo{booktitle}{\emph{Thirty-Sixth
  Conference on Neural Information Processing Systems (NeurIPS)}}.
\newblock
\urldef\tempurl%
\url{https://openreview.net/forum?id=lTZBRxm2q5}
\showURL{%
\tempurl}


\bibitem[Vaswani et~al\mbox{.}(2017)]%
        {attention_is_all_you_need}
\bibfield{author}{\bibinfo{person}{Ashish Vaswani}, \bibinfo{person}{Noam
  Shazeer}, \bibinfo{person}{Niki Parmar}, \bibinfo{person}{Jakob Uszkoreit},
  \bibinfo{person}{Llion Jones}, \bibinfo{person}{Aidan~N. Gomez},
  \bibinfo{person}{Lukasz Kaiser}, {and} \bibinfo{person}{Illia Polosukhin}.}
  \bibinfo{year}{2017}\natexlab{}.
\newblock \showarticletitle{Attention is All you Need}. In
  \bibinfo{booktitle}{\emph{NeurIPS}}.
\newblock


\bibitem[Wen et~al\mbox{.}(2023)]%
        {wen2023transformers}
\bibfield{author}{\bibinfo{person}{Qingsong Wen}, \bibinfo{person}{Tian Zhou},
  \bibinfo{person}{Chaoli Zhang}, \bibinfo{person}{Weiqi Chen},
  \bibinfo{person}{Ziqing Ma}, \bibinfo{person}{Junchi Yan}, {and}
  \bibinfo{person}{Liang Sun}.} \bibinfo{year}{2023}\natexlab{}.
\newblock \bibinfo{title}{Transformers in Time Series: A Survey}.
\newblock
\newblock
\showeprint[arxiv]{2202.07125}~[cs.LG]


\bibitem[White and Hull(1993)]%
        {hull_white}
\bibfield{author}{\bibinfo{person}{Alan White} {and} \bibinfo{person}{John
  Hull}.} \bibinfo{year}{1993}\natexlab{}.
\newblock \showarticletitle{One-Factor Interest-Rate Models and the Valuation
  of Interest-Rate Derivative Securities}.
\newblock \bibinfo{journal}{\emph{Journal of Financial and Quantitative
  Analysis}}  \bibinfo{volume}{28} (\bibinfo{date}{06} \bibinfo{year}{1993}),
  \bibinfo{pages}{235--254}.
\newblock
\urldef\tempurl%
\url{https://doi.org/10.2307/2331288}
\showDOI{\tempurl}


\bibitem[Wu et~al\mbox{.}(2022)]%
        {wu2022autoformer}
\bibfield{author}{\bibinfo{person}{Haixu Wu}, \bibinfo{person}{Jiehui Xu},
  \bibinfo{person}{Jianmin Wang}, {and} \bibinfo{person}{Mingsheng Long}.}
  \bibinfo{year}{2022}\natexlab{}.
\newblock \bibinfo{title}{Autoformer: Decomposition Transformers with
  Auto-Correlation for Long-Term Series Forecasting}.
\newblock
\newblock
\showeprint[arxiv]{2106.13008}~[cs.LG]


\bibitem[Xu(2006)]%
        {convex_risk_measure}
\bibfield{author}{\bibinfo{person}{Mingxin Xu}.}
  \bibinfo{year}{2006}\natexlab{}.
\newblock \showarticletitle{Risk Measure Pricing and Hedging in Incomplete
  Markets}.
\newblock \bibinfo{journal}{\emph{Annals of Finance}}  \bibinfo{volume}{2}
  (\bibinfo{date}{02} \bibinfo{year}{2006}), \bibinfo{pages}{51--71}.
\newblock
\urldef\tempurl%
\url{https://doi.org/10.1007/s10436-005-0023-x}
\showDOI{\tempurl}


\bibitem[Zeng et~al\mbox{.}(2023)]%
        {Zeng2022AreTE}
\bibfield{author}{\bibinfo{person}{Ailing Zeng}, \bibinfo{person}{Muxi Chen},
  \bibinfo{person}{Lei Zhang}, {and} \bibinfo{person}{Qiang Xu}.}
  \bibinfo{year}{2023}\natexlab{}.
\newblock \showarticletitle{Are Transformers Effective for Time Series
  Forecasting?}
\newblock \bibinfo{journal}{\emph{Proceedings of the AAAI Conference on
  Artificial Intelligence}}.
\newblock


\bibitem[Zhou et~al\mbox{.}(2021)]%
        {zhou2021informer}
\bibfield{author}{\bibinfo{person}{Haoyi Zhou}, \bibinfo{person}{Shanghang
  Zhang}, \bibinfo{person}{Jieqi Peng}, \bibinfo{person}{Shuai Zhang},
  \bibinfo{person}{Jianxin Li}, \bibinfo{person}{Hui Xiong}, {and}
  \bibinfo{person}{Wancai Zhang}.} \bibinfo{year}{2021}\natexlab{}.
\newblock \bibinfo{title}{Informer: Beyond Efficient Transformer for Long
  Sequence Time-Series Forecasting}.
\newblock
\newblock
\showeprint[arxiv]{2012.07436}~[cs.LG]


\bibitem[Zhu and Diao(2023)]%
        {zhu2023stochastic}
\bibfield{author}{\bibinfo{person}{Qinwen Zhu} {and} \bibinfo{person}{Xundi
  Diao}.} \bibinfo{year}{2023}\natexlab{}.
\newblock \showarticletitle{From Stochastic to Rough Volatility: A New Deep
  Learning Perspective on Hedging}.
\newblock \bibinfo{journal}{\emph{Fractal and Fractional}} \bibinfo{volume}{7},
  \bibinfo{number}{3} (\bibinfo{year}{2023}), \bibinfo{pages}{225}.
\newblock


\end{thebibliography}
